%% file: main.tex
\documentclass[10pt,twocolumn,letterpaper]{article}

\usepackage{iccv}
\usepackage{times}
\usepackage{epsfig}
\usepackage{graphicx}
\usepackage{amsmath}
\usepackage{amssymb}

\usepackage{multirow}
\usepackage{booktabs}
\usepackage{color}
\usepackage{rotating}
\usepackage{subcaption}
\usepackage{tabularx}
\usepackage[accsupp]{axessibility}

\DeclareMathOperator*{\argmin}{argmin}

\makeatletter
\DeclareRobustCommand\onedot{\futurelet\@let@token\@onedot}
\def\@onedot{\ifx\@let@token.\else.\null\fi\xspace}
\def\eg{\emph{e.g}\onedot} 
\def\ie{\emph{i.e}\onedot} 
\def\cf{\emph{c.f}\onedot}

\newcommand{\B}{\bfseries}
\newcommand\rurl[1]{%
  \href{https://#1}{\nolinkurl{#1}}%
}
\makeatother

\usepackage[pagebackref=true,breaklinks=true,colorlinks,bookmarks=false]{hyperref}
\usepackage{cleveref}

\iccvfinalcopy % *** Uncomment this line for the final submission

\def\iccvPaperID{5693} % *** Enter the ICCV Paper ID here

\ificcvfinal\fi

\begin{document}

%%%%%%%%% TITLE

\title{ViewNet: Unsupervised Viewpoint Estimation from Conditional Generation}

\author{Octave Mariotti
\qquad
Oisin Mac Aodha
\qquad
Hakan Bilen \\[0.1in]
School of Informatics, University of Edinburgh\\
\small{\rurl{groups.inf.ed.ac.uk/vico/ViewNet}}
}

\maketitle
\ificcvfinal\thispagestyle{empty}\fi

\begin{abstract}
Understanding the 3D world without supervision is currently a major challenge in computer vision as the annotations required to supervise deep networks for tasks in this domain are expensive to obtain on a large scale. 
In this paper, we address the problem of unsupervised viewpoint estimation. 
We formulate this as a self-supervised learning task, where image reconstruction provides the supervision needed to predict the camera viewpoint. 
Specifically, we make use of pairs of images of the same object at training time, from unknown viewpoints, to self-supervise training by combining the viewpoint information from one image with the appearance information from the other. 
We demonstrate that using a perspective spatial transformer allows efficient viewpoint learning, outperforming existing unsupervised approaches on synthetic data, and obtains competitive results on the challenging PASCAL3D+ dataset.
\end{abstract}
\vspace{-5pt}

\section{Introduction}
\label{sec:intro}
\input{intro.tex}

\section{Related work}
\label{sec:relwork}
\input{relwork}

\section{Method}
\label{sec:method}
\input{method}

\section{Experiments}
\label{sec:experiments}
\input{experiments}

\section{Conclusion}
\label{sec:conclusion}
\input{conclusion}

\paragraph{Acknowledgments.} OM is supported by Toyota Motor Europe, and HB by EPSRC Visual AI 
grant EP/T028572/1.
\clearpage

{\small
\bibliographystyle{ieee_fullname}
\bibliography{egbib}
}

\onecolumn 
\newpage
\appendix

\input{supplementary}

\end{document}

%% file: intro.tex
% !TEX root = cvpr.tex

Object viewpoint estimation is one of the key components required to enable autonomous systems to understand the 3D world.
Earlier methods~\cite{hinterstoisser2011multimodal, hinterstoisser2012model, brachmann2014learning} were successfully demonstrated to work in controlled environments.
While more recent work, benefiting from modern learnable representations~\cite{tulsiani2015viewpoints, grabner20183d, liao2019spherical}, have been shown to help other vision tasks such as object detection and 3D reconstruction \cite{kundu20183d} and have been deployed in various applications \cite{ranjan2017hyperface,ge2018real,lin2017recurrent}. 

\begin{figure}[t]
    \centering
    \includegraphics[width=0.92\linewidth]{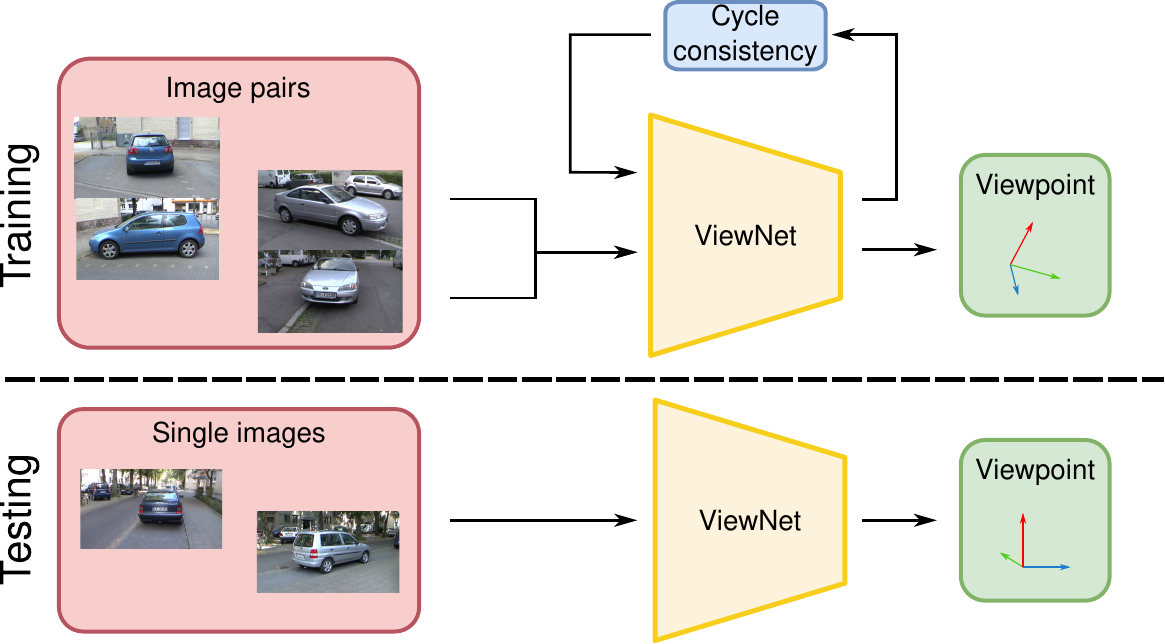}
    \vspace{-5pt}
    \caption{ViewNet learns to extract the camera viewpoint via self-supervised training on a collection of image pairs. At inference it can estimate viewpoint from a single image.
    }
    \vspace{-12pt}
    \label{fig:overview}
\end{figure}

In this paper, we focus on recovering the 3D pose of an object relative to the camera viewing it from a single image. 
This requires an accurate understanding of the 3D structure of the objects depicted in an image, and can be complicated by challenges such as object symmetry.
While large-scale image datasets annotated with semantic supervision (\eg \cite{deng2009imagenet,lin2014microsoft}) have been a key enabler for modern deep networks, obtaining viewpoint annotations can be extremely laborious, expensive, and error-prone.
For instance, the annotation procedure for the commonly used object viewpoint benchmark, PASCAL3D+~\cite{xiang2014beyond}, required annotators to manually select an appropriate 3D CAD model from a pool of models and click on the 2D locations for a set of predefined landmarks for each object instance.
Similarly, creating 3D face pose estimation datasets such as \cite{zhu2017face} involves a time consuming  step of fitting morphable 3D face models to 2D face images.
As a result, creating large-scale viewpoint datasets for a diverse set of objects, especially when 3D CAD models are unavailable, is challenging using existing annotation pipelines.

To overcome this problem, we propose a self-supervised object viewpoint estimation method, \emph{ViewNet}, that learns from an unlabeled (\ie without ground truth pose) collection of image pairs from a given object category  (see Figure~\ref{fig:overview}). % (\eg planes, cars, chairs). 
Our method exploits multi-view consistency and does not require any manual viewpoint annotations. 
Our work is inspired by the analysis by synthesis and conditional generation paradigms~\cite{yan2016perspective, tulsiani2018multi, mariotti2020semi, mustikovela2020self} where we learn to disentangle viewpoint and 3D appearance of objects and reconstruct images based on these disentangled factors.
To this end, we leverage the information contained in image pairs of the same object with different viewpoints.
Such pairs may be generated synthetically or obtained from videos.
Given such an image pair, our method extracts 3D appearance from the first image and viewpoint from the second one, and reconstructs the second one based on this factorization. 
At test time, our method can be used to predict the viewpoint relative to the camera of objects from single images. 
Unlike previous work~\cite{tulsiani2017multi,insafutdinov2018unsupervised}, we are able to leverage supervision directly from pixel values, allowing for more efficient supervision, as well as enabling the generation of images from new viewpoints.

Our main contributions are:
(i) A new conditional generation approach for estimating viewpoint from single images via self-supervision. 
    Our model encodes strong geometric consistency, enabling it to accurately generate novel views that can be used to refine its own predictions. 
(ii) A detailed evaluation of our method on ShapeNet and PASCAL3D+, where we outperform related approaches.
(iii) We highlight the limitations of current evaluation procedures by  showing that a large portion of object instances in certain categories are captured from similar viewpoints rendering their evaluation biased. We also show that the calibration step commonly used to align estimated unsupervised viewpoints with the ground-truth  introduces additional biases. 

%% file: relwork.tex
% !TEX root = cvpr.tex

\noindent{\bf Supervised pose estimation.} 
Effective pose estimation from images has many real-world applications, \eg in robotics or autonomous vehicles, and thus has been extensively studied.
While early works performed pose estimation by matching low-level image descriptors \cite{hinterstoisser2011multimodal, hinterstoisser2012model, brachmann2014learning}, more recent approaches employ deep networks for predicting 3D bounding boxes \cite{rad2017bb8,tekin2018real,grabner20183d}, or classifying viewpoints directly \cite{tulsiani2015viewpoints, kehl2017ssd}. 
Keypoint prediction is a closely related task, and multitask setups where both are jointly learned have been shown to be  successful \cite{tulsiani2015viewpoints, zhou2018starmap}. 
Alternatively, one can be used to learn the other, as one can recover pose by aligning keypoints~\cite{pavlakos20176dof}, or discover them by enforcing a pose-aware sparse representation of objects~\cite{suwajanakorn2018discovery}

Recent work has proposed modeling the topology of the viewpoint space by quantifying the uncertainty with a von~Misses distribution~\cite{prokudin2018deep}, learning 2D image embeddings that are equivariant to 3D pose~\cite{esteves2018cross}, employing a spherical exponential mapping at the regression output~\cite{liao2019spherical}, or introducing cylindrical convolutions~\cite{joung2020cylindrical}.
However, all of these approaches are supervised and requires pose annotations from datasets such as PASCAL3D+~\cite{xiang2014beyond} or LINEMOD~\cite{hinterstoisser2012model}. 
The first of which was manually annotated, and the second was created in a controlled lab setup where poses were collected with each image.
Alternatively, coarse viewpoint estimation can be obtained without manual annotations using structure from motion algorithms on videos~\cite{sedaghat2015unsupervised, novotny2017learning}.
Ground truth pose annotations are challenging to acquire, and recent benchmarks still require human intervention in order to set the coordinate system for each instance and to correct automatic pose errors~\cite{objectron2020}.

\noindent{\bf 3D-aware representations.} 
A parallel line of work learns representations that are aware of the underlying 3D structure of objects from images.
Earlier works employ auto-encoders to disentangle pose and object appearance, with \cite{worrall2017interpretable} or without \cite{kulkarni2015deep} pose supervision. 
More recent works extend this disentangled pose learning from in-plane rotations to full 3D poses by crafting models that reason with spherical representations \cite{cohen2018spherical, esteves20173d}, apply 3D rotations on embeddings to reconstruct images from a different viewpoint \cite{rhodin2018unsupervised}, use denoising auto-encoders to better extract viewpoint information \cite{sundermeyer2018implicit}, or by generalizing variational auto-encoders to spherical functions \cite{falorsi2018explorations}.

First proposed for 2D feature maps, spatial transformers~\cite{jaderberg2015spatial} provide a way to apply in-plane transformations to any representation using spatial resampling and were later extended to 3D convolutions \cite{yan2016perspective}. 
These sampling operations can be used to represent complete 3d scenes from multiple views~\cite{sitzmann2019deepvoxels}.
In a related analysis by synthesis approach, \cite{chen2020category} also learn pose representations via an appearance based reconstruction loss. 
At inference time, they iteratively optimize for the viewpoint that minimizes the appearance loss between the synthesized view and the input image.
However, apart from a few unrealistically simplified experiments, all of these methods require 3D annotations in order to learn meaningful embeddings. 
Unlike in-plane rotations, which are simple enough to learn in an unsupervised way, 3D rotations can cause drastic appearance changes that are often too complex for networks to learn without pose annotations~\cite{mariotti2020semi}.

\noindent{\bf Viewpoint-conditioned generation.} 
An increasingly popular way of learning interpretable representations is by using a generation process conditioned on the relevant information.
The two main ways of building such representation rely either on encode-decoder approaches, using image pairs where semantics are shared \cite{whitney2016understanding, jakab2018conditional, mariotti2020semi}, or on adversarial models to generate new samples in a controllable way \cite{chen2016infogan, nguyen2019hologan}. 
Both techniques have been shown to estimate viewpoints without labels \cite{tulsiani2018multi, insafutdinov2018unsupervised, mustikovela2020self}. 
Encode-decoder approaches are closely related to the field of unsupervised 3D reconstruction \cite{henzler2019escaping, niemeyer2020differentiable, oechsle2019texture, olszewski2019transformable}.
In contrast to \cite{tulsiani2018multi, insafutdinov2018unsupervised} that do both 3D reconstruction and pose estimation, we propose a simpler fully self-supervised approach that is able to leverage appearance matching as supervision, allowing for novel view synthesis that can be used to further refine predictions.

SSV~\cite{mustikovela2020self} uses an adversarial model to generate objects with random rotations while learning to regress viewpoint at the same time. 
In contrast, our proposed method ensures geometric consistency during the image generation process, allowing for more robust viewpoint estimation.
Furthermore, GAN training can be unstable~\cite{metz2016unrolled, arjovsky2017towards}, an issue often reflected in the auxiliary objectives required to guide training. 
In contrast, our method operates via image reconstruction alone, and can easily generate images from novel viewpoints.
Several non-adversarial generative approaches have also been proposed~\cite{mariotti2020semi, chen2020category} that reconstruct specific object instances in order to leverage pixel-level supervision. 
However, unlike our approach, these methods require at least a partially labeled training set.

%% file: method.tex
% !TEX root = cvpr.tex

Given a collection of unlabeled images $\mathcal{T}$, at training time we aim to learn a function $f^v:\mathcal{I}\rightarrow\mathcal{V}$ that can map from image space $\mathcal{I}$ to pose space $\mathcal{V}$. 
At test time, we can then apply this function to a single image $I$, containing an object of interest, in order to estimate its 3D viewpoint $v$ relative to the camera.
3D viewpoints can be represented in different ways, including the Euler angles (azimuth, elevation and tilt), or with a rotation matrix $R \in SO_3$, and we use both representations interchangeably.

\begin{figure*}[t]
    \centering
    \includegraphics[width=0.9\linewidth]{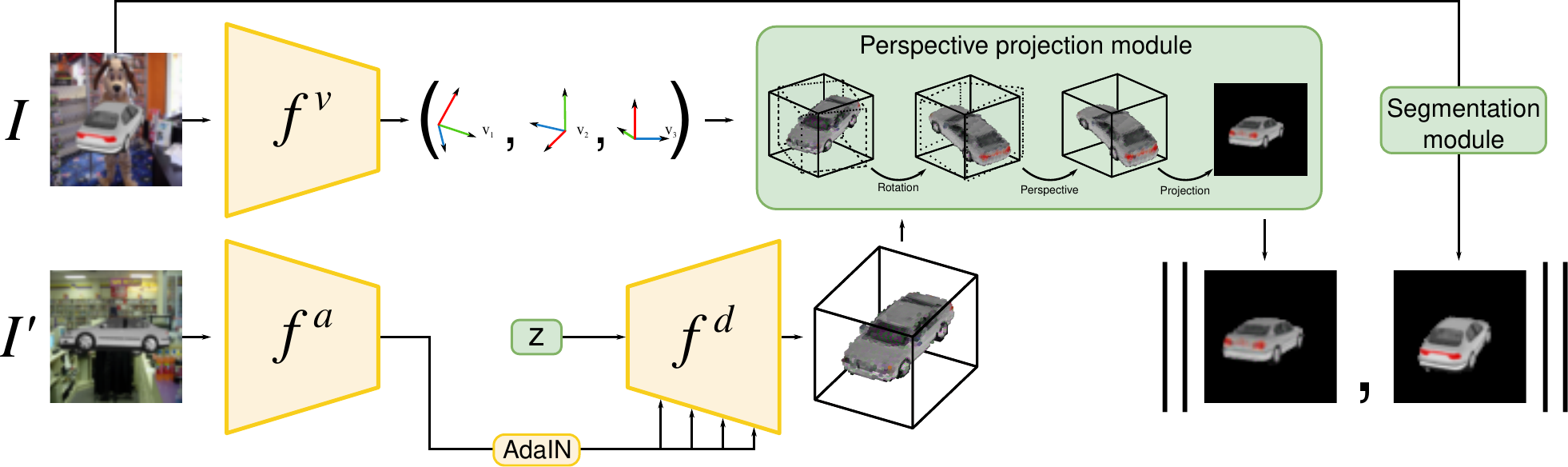}
    \vspace{-5pt}
    \caption{Overview of ViewNet. $f^v$ is the viewpoint prediction network. 
    At training time, $f^a$ encodes the object appearance embedding from image $I'$ which is decoded by $f^d$ into a 3D representation and transformed by the estimated viewpoint into an image in the same pose as $I$ using the projection module. 
    This reconstruction, which can be segmented, is then compared to $I$ to guide training.
    Yellow blocks indicate learned parameters, while green ones are fixed or analytical modules.
    }
    \vspace{-13pt}
    \label{fig:arch}
\end{figure*}

As ground-truth viewpoints of the objects in $\mathcal{T}$ are challenging to acquire, we formulate our problem as a self-supervised task that uses principles from conditional generation and synthesis by analysis.
To this end, we propose to factorize the viewpoint and appearance of objects via two functions $f^v$ and $f^a$. 
Given an image $I$, $f^a$ outputs an appearance feature $\mathbf{a}$ for the object contained in it.
The decoder $f^d$, can reconstruct the image $I$ given the pose of the object $\mathbf{v}$ and its appearance $\mathbf{a}$. 
$f^v$, $f^a$, and $f^d$ are instantiated as neural networks parameterized by $\theta^v$, $\theta^a$ and $\theta^d$ respectively.
Clearly such a factorization is not guaranteed without some constraints on $f^v$ and $f^a$.
To overcome this ambiguity, we use image pairs of rigid objects at training time that differ by their viewpoint.
Such pairs can be extracted from video sequences, generated by perturbing still images, or rendered from 3D CAD models. 
Hence we assume that the set of unlabeled images $\mathcal{T}$ can be described as $N$ image pairs $\mathcal{T}=\{(I_i,I'_i)\}_{i=1}^N$ where each pair contains images of the same object instance from two different viewpoints $(v_i,v'_i)$, where the actual viewpoint information, relative or absolute, is not available.
Given an image pair $(I_i,I'_i)$, we propose to extract pose features $\mathbf{v}$ from $I_i$ and appearance features $\mathbf{a'}$ from $I_i'$, and use them to reconstruct $I_i$. 
An overview of our model is shown in Figure~\ref{fig:arch}. 
Our learning task consists of solving the following objective:
\begin{equation}
    \min_{\theta^v, \theta^a, \theta^d} \sum_{(I,I') \in \mathcal{T}} \lvert\lvert f^d(f^a(I'), f^v(I)) - I \rvert\rvert.
\end{equation}

\subsection{Pose estimation network $f^v$}
\label{subsec:pose_prediction}

We design the pose estimation network to output a point on the 3D unit sphere (\ie $f^v(I) = \mathbf{v} \in S^2$), and uniquely map each point on the sphere to a viewpoint. 
To this end, we apply an orthogonalization operation to $f^v$'s output with the following steps. 
First we define an arbitrary vector $\mathbf{u} \in S^2$ that represents the upwards direction, then we apply two successive cross products, $\mathbf{w} = \mathbf{v} \times \mathbf{u}$ and $\mathbf{u'} = \mathbf{w} \times \mathbf{v}$, and normalize the results to obtain orthogonal vectors.
Finally we define the rotation matrix $R$ as $ \left[ \mathbf{v, w, u'}\right]$. 
This matrix is then used to rotate the object representation during the generative stage, described later. 
This approach uses an arbitrarily chosen upwards direction, meaning we assume images do not contain in-plane rotations.
However, in the more general case, $\mathbf{u}$ can be learned jointly with $\mathbf{v}$, effectively describing the full range of 3D rotations.

The main pitfall of unsupervised viewpoint estimation is the collapse of predictions caused by symmetries. 
Current approaches work well on simple objects \eg a cube with each face colored differently. 
However, real world objects tend to have at least one, if not multiple, symmetric viewpoint pairs. 
We say that two viewpoints $v, v'$ form a symmetric pair, $v\sim v'$, if the image produced by observing the object from $v$ is close to that produced from $v'$. For instance, in most cars, $(a, e, t) \sim (a + \pi, e, t)$ form a symmetric pair for any azimuth $a$, elevation $e$, and camera tilt $t$. 
As a results of this, unsupervised methods based on reconstruction often equate those two viewpoints, leading to a collapse of the predictions. 
Different workarounds have been proposed to mitigate this, such as using adversarial losses to enforce a prior on the pose distribution \cite{tulsiani2018multi}, using multiple prediction heads \cite{tulsiani2018multi, insafutdinov2018unsupervised}, or enforcing some symmetric consistency in the predictions using a flipped version of the image \cite{mustikovela2020self}. 
The main drawback of this last approach
is that it is only valid for a left-right planar symmetry, and would likely fail in the aforementioned car example. 
To overcome this problem, we use multiple prediction heads for our pose estimator, resulting in multiple hypotheses for $\mathbf{v}$.
Each head can learn to specialize in a subset of the viewpoints, and in the case of a symmetric pair $v \sim v'$, both can simultaneously be predicted by two different heads.

In practice, each predictor head $f^v_m$ outputs a viewpoint prediction, and the one associated with the lowest reconstruction error is chosen as the prediction at training time:
\begin{equation}
    \mathbf{v^*}=f^v_{m^*}(I) \text{~~s.t.~~} m^*=\argmin_{m \in M}\lvert\lvert f^d(f^a(I'), f^v_m(I)) - I \rvert\rvert,
    \label{eq:mstar}
\end{equation}
where $f^v_m$ denotes $m$-th viewpoint predictor and $M$ is the number of heads.
Gradients will only be propagated through $m^*$, ensuring that symmetric pairs get separated.

At test time, ViewNet only requires the pose prediction network $f^v$, and does not need $f^a$ or $f^d$ in order to make a prediction. 
To achieve this, we jointly train a selection head, which is tasked with picking the best prediction for each input image given the range of options. 
The task is to minimize the cross-entropy between the selection prediction and a one-hot distribution representing $m^*$, computed via Eqn.~\ref{eq:mstar}. 
Although $m^*$ is not guaranteed to be the prediction closest to ground truth pose, we observe it is enough to differentiate between symmetric viewpoint pairs.
Compared with~\cite{insafutdinov2018unsupervised}, this allows us to efficiently maintain multiple hypotheses at test time, which translates to more robust predictions, and we do not require complex solutions such as reinforcement learning as in~\cite{tulsiani2018multi}.
\subsection{Appearance encoding network $f^a$}
The appearance $f^a(I') = \mathbf{a'} \in \mathbb{R}^n$ of the object represented in the input image is also learned with a convolutional network. 
In a standard encoder-decoder architecture, $\mathbf{a'}$ would be used as an input to $f^d$ to produce a reconstruction. 
However, this offers no guarantee that the viewpoint  $\mathbf{v'}$ and appearance $\mathbf{a'}$ embeddings are correctly factorized. 
In particular, information about $\mathbf{v'}$ could be encoded in $\mathbf{a'}$. 
This means that a change in $\mathbf{v'}$ could induce changes in the appearance of the reconstruction. 
In extreme cases, the network could even ignore $\mathbf{v}$ and reconstruct $I$ by memorizing the $(I, I')$ pairs. 
To mitigate this, we use an object-conditional generation process which makes use of adaptive instance normalization (AdaIN)~\cite{huang2017arbitrary}. 
Initially developed for style transfer, this approach is popular in GANs~\cite{brock2018large, nguyen2019hologan, mustikovela2020self} due to its ability to adapt the generation process at different scales. 

Our generation pipeline works by refining a random static code $\mathbf{z} \in \mathbb{R}^m$ through the decoder network to the final rendering stage. 
$\mathbf{z}$ is randomly picked from a normal distribution at the beginning of the training process, and remains constant. 
Its purpose is to encode the average object in a canonical pose. 
The appearance of the object is gradually encoded by AdaIN layers (see Figure~\ref{fig:arch}), which apply an affine transformation to the features parameterized by $\mathbf{a}$. 
As they are applied uniformly over each feature channels, it is impossible for them to alter local information of the features.
Additionally, a standard encoder-decoder architecture would only use $\mathbf{a}$ as input of the decoder, meaning fine details of the object can be lost during the complex decoding process. By comparison, applying transformations across different layers means they can influence all levels of the reconstruction, resulting in more faithful reconstructions.

\subsection{Decoder network $f^d$}
\label{subsec:rendering}

In order to ensure an accurate viewpoint prediction, we aim to strictly enforce geometric consistency during the generation process. 
To this end, $f^d$ is modeled using 3D convolutional layers, and uses a 3D spatial transformer with perspective for image rendering, similar to those used in~\cite{yan2016perspective}, and is combined with a pseudo-ray tracing operation inspired by~\cite{tulsiani2017multi}.  
Placing the rotation at the final stage of the network, as close as possible to the reconstruction loss, ensures that gradients are efficiently propagated to $f^v$. 
The absence of parametric transformations between $f^v$ and the target image guarantees that viewpoint errors cannot be compensated for by convolutional layers, as can happen in GAN-based models.

Our rendering module consists of three main steps and is related to those used in~\cite{yan2016perspective, tulsiani2017multi, tulsiani2018multi, insafutdinov2018unsupervised},  however our pipeline also makes use of texture information. 
Specifically, the steps involve: (i) Rotation. Given a 3D volume $V$, a spatial transformer can be used to rotate it using a rotation matrix $R$. The new volume is obtained by resampling the data along the rotated axis.
(ii) Perspective. Similarly, perspective can be simulated with a spatial transformer. The single point perspective of a pinhole camera will have the effect of decreasing the apparent size of objects proportional to distance. We can therefore resample the volume by dilating close points and contracting distant ones.
(iii) Projection-based ray-tracing. Finally, the volume is projected to a two-dimensional image plane. 
As parts of the objects will be subject to self-occlusion, we use a pseudo ray-tracing operation to compute which voxels will appear in the output image, ensuring geometric consistency.

For each entry in the 3D volume $V$, the first three channels $C$ represent the RGB channels of an image, while the fourth one $Q$ is an occupancy map, containing  information about the shape of the object. 
The value of each cell is interpreted as the probability of the object occupying the corresponding spatial location. 
To compute the projection, we have to estimate where each light ray is likely to stop. Since we already accounted for the perspective, all our rays are parallel, leaving only the depth of each stopping point to be computed. 
Compared with \cite{tulsiani2017multi}, we do not have to compute a path for each light ray \ie it is embedded in the shape of the tensor. 
This means we can compute all lights paths simultaneously using efficient parallel operations, in a manner similar to the orthographic projection used in~\cite{gadelha20173d}. 
The probability of the light ray, at pixel coordinates $i,j$, stopping at depth $k$ is given by
\begin{equation}
    Q'_{i,j,k} =  Q_{i,j,k} \times \prod_{l=0}^{k-1} (1-Q_{i,j,l}),
\end{equation} with the convention that an empty product is equal to 1. 
 The first term represents the probability of the voxel at coordinates $(i,j,k)$ being occupied, and the second one is the probability of all the previous ones being not visible. 
 Hence, the final pixel value at coordinate $i,j$ is
\begin{equation}
    \hat{I}_{i,j} = \sum_{k=1}^{n}\left[ C_{i,j,k} \times Q_{i,j,k} \times \prod_{l=0}^{k-1} (1-Q_{i,j,l})\right].
\end{equation}
This is similar to the formulation in~\cite{tulsiani2017multi, tulsiani2018multi}, although in our case, the ray-tracing is parallelized and used to sample RGB values, rather than computing depth or ray termination.

A failure case of our approach consists of ViewNet using the volume $V$ as a canvas and ``painting'' the object in different poses on the sides. 
More generally, this results in errors in the predicted shape of the object, since we do not know which pixels belong to it. 
To address this, instead of trying to directly estimate occupancy $Q$, we learn $Q'$ such that $Q = S + Q'$ where $S$ is a three dimensional Gaussian distribution centered on $V$. 
$Q'$ can be interpreted as a residual that deforms a shape prior $S$ so that it matches the shape of the observed object.
$S$ encodes a prior for the shape and position of the object, while discouraging the network from using voxels that are far away from the center of the volume. 

\subsection{Cycle consistency supervision}
Using appearance supervision, as opposed to only object silhouettes as in~\cite{yan2016perspective, tulsiani2018multi, insafutdinov2018unsupervised}, enables ViewNet to also represent appearance information.
This has two key advantages.
First, our method can generate images of objects from novel views.
Second, we can use these novel views to regularize our model during training by enforcing consistency between a generated image and its known viewpoint.

Given a randomly sampled viewpoint $\mathbf{\tilde{v}} \sim \mathcal{U}(\mathcal{V})$, we can render a novel image $\widetilde{I} = f^d(\mathbf{\tilde{v}, a'})$ using appearance information in $\mathbf{a'}$  extracted from image $I'$. 
By feeding this to $f^v$, we can compute the distance between the sampled viewpoint $\tilde{v}$ and its estimated viewpoint $f^v(\widetilde{I})$, \ie $\mathcal{L}_{cycle} = \lvert\lvert f^v(\widetilde{I}) - \mathbf{\tilde{v}}\rvert\rvert$ and backpropagate this error to the viewpoint estimator.
Assuming the reconstructions are of sufficient quality, this allows us to generalize beyond the potentially limited set of poses that are present in the training set, and these newly generated samples help regularize the viewpoint estimation network.

%% file: experiments.tex
% !TEX root = main.tex

\begin{figure*}[t]
    \centering
    \begin{subfigure}[b]{0.22\textwidth}
        \centering
        \includegraphics[width=\textwidth]{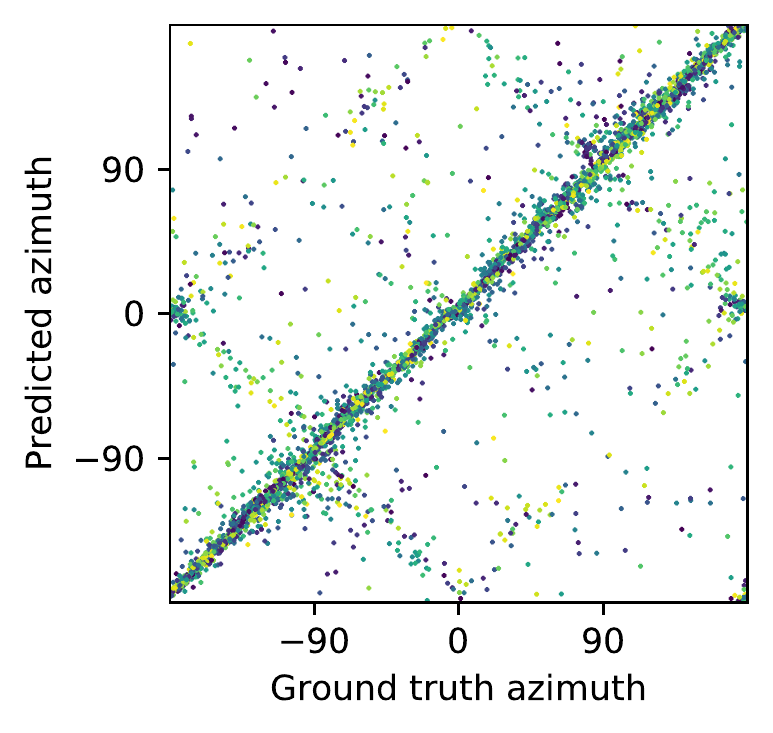}
        \caption{aeroplane}
    \end{subfigure}
    \hfill
    \begin{subfigure}[b]{0.22\textwidth}
        \centering
        \includegraphics[width=\textwidth]{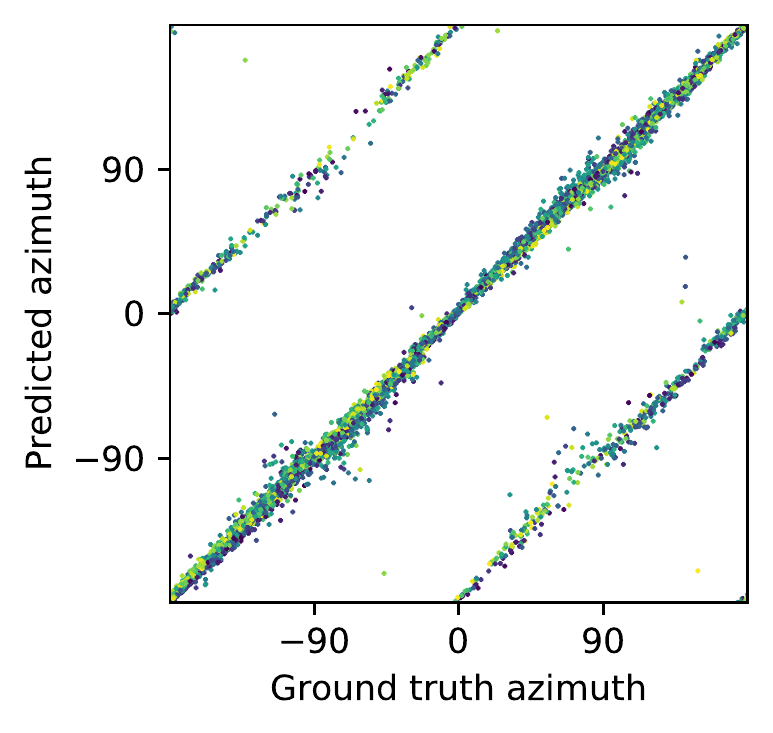}
        \caption{car}
    \end{subfigure}
    \hfill
    \begin{subfigure}[b]{0.22\textwidth}
        \centering
        \includegraphics[width=\textwidth]{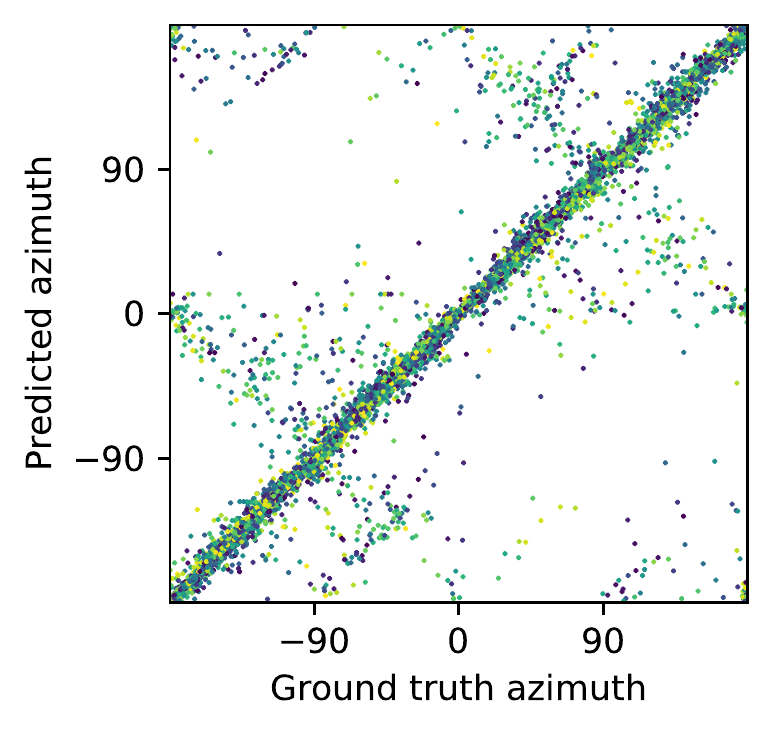}
        \caption{chair}
    \end{subfigure}
    \begin{subfigure}[b]{0.30\textwidth}
        \centering
        \includegraphics[width=\textwidth]{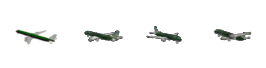}
        \includegraphics[width=\textwidth]{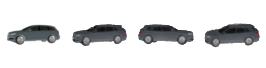}
        \includegraphics[width=\textwidth]{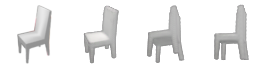}
        \caption{Reconstruction proposals}
    \end{subfigure}
    \vspace{-5pt}
    \caption{(a) - (c) Comparison of ground truth versus predicted azimuth on three ShapeNet categories. A perfect predictor would appear as a single diagonal line.
    (d) Candidate reconstructions for each head. The left image is the input of the pose estimator, and the three following images are the renderings for each head, ranked by increasing reconstruction error.
    }
    \vspace{-5pt}
    \label{fig:comp_shapenet}
\end{figure*}

Here we present 3D pose estimation results on both synthetic and real image datasets.

\subsection{Implementation details}
ViewNet consists of three sub-networks: $f^v$, $f^a$, and $f^d$.
Both $f^v$ and $f^a$ contain seven convolutional layers interleaved with batch normalization and ReLU activation functions respectively.
$f^v$ takes a $64\times 64$ RGB image $I$ as input and outputs $M=3$ viewpoint hypotheses. 
$f^a$ encodes a second RGB image $I'$, depicting the same object instance captured from another viewpoint, and outputs a $256$ dimensional appearance vector.
The input to $f^d$ is a $1024$ dimensional fixed canonical code vector. 
The canonical code is passed through seven 3D transposed convolutions, each followed by a ReLU, and the feature maps are further conditioned on the output of $f^a$ via adaptive instance normalization (AdaIN) layers.
The resulting 3D feature map is projected to an image based on the predicted pose and used to compute the reconstruction error w.r.t. $I$.  
We use a perceptual loss~\cite{johnson2016perceptual}, as it provides more informative gradients compared with standard pixel-level reconstruction losses.
The supplementary material provides additional details.

In all experiments, we set the minibatch size to 64 and use the Adam optimizer~\cite{kingma2014adam}. 
For each experiment, we train a separate model per category, and select the model with the best performances on a held out validation set, stopping the training if no improvement is observed for 30 epochs.
As our method is unsupervised, all viewpoints are predicted up to a random rotation. 
In order to evaluate our model, we must align its predictions with the ground truth. 
The standard alignment technique, performed by~\cite{tulsiani2018multi,insafutdinov2018unsupervised}, involves computing the rotation that best aligns the predicted viewpoints with the ground truth. 
This is obtained from a small batch of validation images, using the orthogonal Procrustes algorithm. 
An alternative alignment procedure, used in~\cite{mustikovela2020self}, learns the parameters of a more flexible affine transformation that best maps the predicted viewpoints and to the ground truth. 
This can shrink and/or expand the predicted viewpoint estimation compared to applying a single 3D rotation to translate them.
We discuss potential issues with this approach in section~\ref{subsec:exp_pascal}.
We report performances in standard viewpoint estimation measures: accuracy at $30^o$ and median angular error. 

\subsection{ShapeNet results}

\begin{table}[t]
\small
\setlength{\tabcolsep}{3pt}
\newcolumntype{R}{>{\raggedleft\arraybackslash}X}
  \centering
  \begin{tabularx}{\linewidth}{lRRR|RRR}
    \toprule
    & \multicolumn{3}{c}{Accuracy (\%) $\uparrow$} &\multicolumn{3}{c}{Median error ($^o$) $\downarrow$}\\
    & airplane & car & chair & airplane & car & chair\\
    \midrule
    MVC \cite{tulsiani2018multi}                   &    69 &    87 &    81  &  14.3 &   5.2 &   7.8\\
    Pointcloud \cite{insafutdinov2018unsupervised} &    75 &    86 &    86  &   8.2 &\B 5.0 &   8.1\\
    ViewNet                                        &    82 &    89 &    89  &   8.6 &   6.7 &   7.3\\
    ViewNet + cycle                                &\B  86 &\B  91 &\B  92  &\B 7.7 &   6.7 &\B 7.0\\
  \bottomrule
  \end{tabularx}
  \vspace{-5pt}
  \caption{ShapeNet results for unsupervised methods. Bold entries are the best performing models for each category. 
  }
  \vspace{-10pt}
  \label{tab:shapenet}
\end{table}

Following~\cite{tulsiani2018multi, insafutdinov2018unsupervised}, we evaluate ViewNet on the ShapeNet dataset~\cite{chang2015shapenet}, which contain 7.5k, 6.8k, and 4k 3D CAD models for cars, chairs, and planes respectively. Training, validation, and testing sets are created by splitting CAD models into (0.7, 0.1, 0.2) fractions respectively.
To render image pairs, we randomly select viewpoints and light sources uniformly over the $\left[0^o, 360^o\right)$ azimuth range and $\left[-20^o, 40^o\right]$ elevation range.
We report results for the standard setting where each CAD model is rendered from five random viewpoints at train and test time.

The results in Table~\ref{tab:shapenet} show that ViewNet outperforms existing unsupervised approaches, except for median error on cars.
ViewNet learns to reconstruct textures in addition to shape, and this supervision is more informative compared to only binary masks, as we can leverage texture cues to efficiently disentangle symmetric viewpoints. 
For example, red tail lights on a car can indicate the rear. 
This penalizes a model that would reconstruct white headlights in their place. 
We investigate this further in our ablation study.

We observe that viewpoint cycle consistency provides a further boost in accuracy (`ViewNet + cycle'). 
Here, novel views are rendered at the same time as the regular reconstruction objective and then fed back to the viewpoint estimator. 
This indicates that our model can generate both novel and accurate images for a  given viewpoint and learns to refine its output in a self-learning manner.

We analyze the viewpoint predictions of ViewNet in Figure~\ref{fig:comp_shapenet}, and plot the predicted azimuth against the ground truth. 
The strong diagonal indicates accurate predictions, while off-diagonal points are errors. 
The results reveal that the majority of the errors are caused by symmetries. 
For example, the car category shows a second line of predictions shifted by $180^o$. 
This corresponds to the $(a,e,t) \sim (a+\pi,e,t)$ symmetry mentioned in section~\ref{subsec:pose_prediction}. 
Other categories showcase different symmetry induced issues, with planes and chairs having a retrograde symmetry $(a,e,t) \sim (\pi-a,e,t)$. 
Samples for each proposed viewpoint are shown in Figure~\ref{fig:comp_shapenet} (d). 
One can see that the global input shape is matched relatively well across the different predicted views.  
Interestingly, reconstruction error is not necessarily directly correlated with viewpoint error, as the second proposed viewpoint for the car has lower reconstruction error than the third, despite being rendered from a completely different viewpoint.

\paragraph{Ablation study.}
In Table~\ref{tab:ablation}, we study the effect of each proposed component in our pipeline.
First, we reduce the number of heads in the viewpoint estimators to one and observe a large overall  drop in the viewpoint accuracy.
For the car category, the single-head estimator cannot deal with the front/back symmetries, resulting in a large performance loss.
Second, we modify ViewNet to reconstruct a binary segmentation mask similar to~\cite{tulsiani2018multi, insafutdinov2018unsupervised}, instead of the pixel values.
Using segmentation masks as targets achieves results comparable with previous segmentation-based approaches in Table~\ref{tab:shapenet}. 
This indicates that ViewNet can leverage texture information to achieve better predictions.
Third,  we remove our Gaussian shape prior and directly estimate the occupancy grid $Q$, instead of $Q'$, and observe that this does not have any significant affect on planes and cars, but causes a dramatic drop for chairs as the network tries to `paint' the object on the faces of the volume.
Next we evaluate the conditioning strategy by removing the AdaIN layers and feeding the output of $f^a$ in the first layer of $f^d$, similar to a traditional encoder-decoder pair. 
While this does not cause drastic performance issues, the reconstructions are less accurate, limiting the ability of this model to use them for self-training.
Finally, we replace the analytic renderer with a learnable decoder using the deconvolutional architecture from~\cite{nguyen2019hologan}. In addition to causing the largest performance drop of all ablations, reconstructions from this model do not exhibit geometric consistency as the generated views do not smoothly change as the object rotates. Qualitative samples are provided in the supplementary materials.

\begin{table}[t]
\small
\setlength{\tabcolsep}{2pt}
\newcolumntype{R}{>{\raggedleft\arraybackslash}X}
  \centering
  \begin{tabularx}{\linewidth}{lRRR|RRR}
    \toprule
   & \multicolumn{3}{c}{Accuracy (\%) $\uparrow$} & \multicolumn{3}{c}{Median error ($^o$) $\downarrow$}\\
   & airplane & car & chair & airplane & car & chair\\
   \midrule
     ViewNet             &\B 82 &\B 89 &\B 89 &\B 8.6 &   6.7 &\B 7.3\\
     Single-head         &   72 &   51 &   66 &  18.1 &  27.1 &  16.3\\
     Segmentation Target &   71 &   85 &   88 &  12.9 &   8.0 &   8.1\\
     No Shape Prior      &   78 &\B 89 &   73 &   9.6 &\B 6.6 &  31.3\\
     Encoder-Decoder     &\B 82 &\B 89 &   88 &   8.7 &   6.8 &   7.8\\
     HoloGAN-Decoder     &   66 &   52 &   72 &  19.6 &  27.6 &  14.5\\
     Constant            &   20 &   22 &   19 &  61.7 &  65.2 &  58.1\\

  \bottomrule
  \end{tabularx}
  \vspace{-5pt}
  \caption{Ablation study results. Here we compare different variants of ViewNet  on ShapeNet.
  }
    \vspace{-15pt}
  \label{tab:ablation}
\end{table}

\subsection{PASCAL3D+ results}
\label{subsec:exp_pascal}

\begin{table*}[t]
\small
\setlength{\tabcolsep}{3pt}
\newcolumntype{R}{>{\raggedleft\arraybackslash}X}
  \centering
  \begin{tabularx}{\textwidth}{cclRRRRRRRRRRRR}
    \toprule
     &  &                         & airplane &   bike &   boat & bottle &    bus &    car &  chair &  table &  mbike &   sofa &  train & tv \\
   \midrule
    \multirow{8}{*}{\begin{sideways}Accuracy ($\%$)\end{sideways}} &
    \multirow{6}{*}{\begin{sideways}\footnotesize Unsupervised\end{sideways}} &
        Constant                       &     45 &     23 &     28 &     96 &     79 &     29 &     58 &     48 &     32 &   81 &     95 &\B   89\\
    & & VGG view*                      &     64 &     63 &     25 &     96 &     78 &     56 &     76 &     48 &     46 &   86 &\B   96 &     85\\
    & & SSV* \cite{mustikovela2020self}&     -- &     -- &     -- &     -- &\B   82 &     67 &     -- &     -- &     -- &   -- &\B   96 &     --\\
    & & ViewNet*                       & \B  72 &     79 &     29 &     96 &     75 &     84 &\B   86 &     52 &     72 &   87 &\B   96 &\B   89\\
    & & ViewNet                        & \B  72 &\B   81 &     38 &\B   97 &     75 &     87 &     82 &     54 &     75 &   86 &     85 &     86\\
    & & ViewNet + cycle *              &     67 &     70 &     23 &     96 &     77 &     87 &     83 &     50 &     74 &\B 89 &     95 &     87\\
    & & ViewNet + cycle                &     71 &     80 &\B   47 &     96 &     80 &\B   88 &     83 &\B   57 &\B   78 &    88 &     88 &     82\\
    \cline{2-15}
    & \multirow{2}{*}{\begin{sideways}\footnotesize Sup.\end{sideways}}
      & Liao \cite{liao2019spherical}  &\B   88 &\B   88 &     61 &\B   96 &\B   97 &     93 &\B   93 &\B   74 &\B   93 &\B   98 &     84 &\B   95\\
    & & Grabner \cite{grabner20183d}   &     83 &     82 &\B   64 &     95 &\B   97 &\B   94 &     80 &     71 &     88 &     87 &\B   93 &     86\\
    
    \midrule
    \multirow{8}{*}{\begin{sideways}Median error ($^o$)\end{sideways}} &
    \multirow{6}{*}{\begin{sideways}\footnotesize Unsupervised\end{sideways}} &
        Constant                       &   32.6 &   56.6 &   61.6 &    8.2 &   16.7 &   55.0 &   25.2 &   31.9 &   53.9 &   13.6 &    8.8 &   14.0\\
    & & VGG view*                      &   20.8 &   22.2 &   55.8 &    7.9 &    9.7 &   25.8 &   14.4 &   29.6 &   33.0 &   10.0 &    8.6 &\B 11.3\\
    & & SSV* \cite{mustikovela2020self}&     -- &     -- &     -- &     -- &\B  9.0 &   10.1 &     -- &     -- &     -- &     -- &\B  5.3 &     --\\
    & & ViewNet*                       &   15.0 &   16.0 &   54.1 &    8.1 &   16.1 &   12.9 &   12.3 &   27.0 &   16.9 &   10.1 &    9.1 &   14.4\\
    & & ViewNet                        &\B 14.0 &   13.4 &   38.4 &\B  7.2 &   16.2 &    5.9 &\B 10.1 &\B 24.4 &   14.2 &\B  9.2 &    7.4 &   13.8\\
    & & ViewNet + cycle *              &   18.2 &   17.1 &   61.3 &    8.0 &   16.3 &    6.7 &   11.7 &   28.8 &   14.7 &\B  9.2 &    9.3 &   14.0\\
    & & ViewNet + cycle                &   14.4 &\B 12.2 &\B 20.6 &\B  7.2 &   14.9 &\B  5.6 &   11.5 &   25.0 &\B 11.6 &   11.5 &   15.6 &   15.8\\
    \cline{2-15}
    & \multirow{2}{*}{\begin{sideways}\footnotesize Sup.\end{sideways}}
      & Liao \cite{liao2019spherical}  &\B  9.2 &\B 11.6 &   20.6 &\B  7.3 &    3.4 &\B  4.8 &\B  8.2 &\B  8.5 &\B 12.1 &\B  8.7 &\B  6.1 &\B 10.1\\
    & & Grabner \cite{grabner20183d}   &   10.0 &   15.6 &\B 19.1 &    8.6 &\B  3.3 &    5.1 &   13.7 &   11.8 &   12.2 &   13.5 &    6.8 &   11.0\\

  \bottomrule
  \end{tabularx}
  \vspace{-5pt}
  \caption{PASCAL3D+ results. Bold entries indicate the best performing models in each category. Entries followed by a star~(*) use a linear regression alignment procedure, and those without use a single global rigid alignment.
  }
  \label{tab:pascal}
    \vspace{-5pt}
\end{table*}

Next we evaluate ViewNet on the challenging real-world PASCAL3D+~\cite{xiang2014beyond} dataset. 
It contains real images from the PASCAL VOC and ImageNet datasets along with annotated viewpoint, including azimuth and elevation.
As this dataset does not provide image pairs that contain the same object instance with varying viewpoints, we evaluate our models with 10-views per CAD model trained on ShapeNet.
As PASCAL3D+ images have backgrounds, we synthetically add random background images from SUN397~\cite{xiao2010sun} to our ShapeNet rendered views during training. 
These backgrounds are only added to the input training pairs to make ViewNet robust to backgrounds at test time. 
However, ViewNet is trained to reconstruct only the object, as it would require additional logic to reconstruct the background.

We report our results in Table~\ref{tab:pascal}.
We observe that for some categories \eg bottle, bus, sofa, train, and tv monitors, the ranges of viewpoints it contains are extremely restricted and concentrated around specific viewpoints.
We reason that the viewpoint alignment procedure used for unsupervised methods is very effective in reaching strong performances on these classes.
To test this hypothesis, we build a simple viewpoint predictor, a constant predictor, that outputs the average viewpoint from the validation set for each object category. This mimics the behavior of an untrained viewpoint estimator that has not learned anything useful and gets calibrated on validation data.
We see that this method performs surprisingly well and even outperforms~\cite{grabner20183d}, a supervised approach on some categories.
Even on non-trivial categories, the constant predictor performs surprisingly well, for instance, it obtains 43\% accuracy on airplanes and 58\% on chairs. By comparison, the same predictor on ShapeNet achieves a much lower performance (see Table~\ref{tab:ablation}), as the dataset was specifically crafted not to be biased.

To mitigate biases in the evaluation set, we propose a different evaluation strategy that consists of splitting the viewpoint space into discrete bins, and  then averaging performance over each bin. Doing so prevents biased predictors from reaching near-perfect performance. Results under this scheme are presented in the supplementary material.

As an additional baseline, we reproduce the setup used in SSV~\cite{mustikovela2020self} and fit a linear regressor to VGG16~\cite{simonyan2014very} Conv5 features, and train it to regress the pose using the same small number of PASCAL3D+ images we use to align our predictions -- see `VGG View' results in Table~\ref{tab:pascal}.

We directly evaluate our ShapeNet trained ViewNet model on PASCAL3D+ images.
We provide results for two alignment methods, the optimal rotation using orthogonal Procrustes, and the linear regression as used in SSV~\cite{mustikovela2020self}, which takes the predicted viewpoint and applies a linear regressor to modify its predictions. 
Depending on the category, two behaviors can be identified: either the two alignment procedures provide similar results (\eg bike, bottle), or the linear regression approach significantly outperforms the optimal rotation. We observe that the second behavior is correlated with categories where PASCAL3D+ contains highly biased viewpoints, \ie where most viewpoints are clustered around a single one. 
We theorize that the linear regression approach can artificially boost performance in those categories by collapsing viewpoint predictions towards the common view. 
This can be achieved by learning zero weights for the predicted viewing angles and encoding the average viewpoint as the bias term.

Similar to our ShapeNet experiments, we also evaluate the impact of training with our cycle based generated views. Depending on the categories, it often provides a small accuracy boost at the cost of higher median error. This median error increase could be due to the higher domain gap between generated views and real-world images.

\subsection{Other dataset results}
Up until this point, we have only trained ViewNet on the synthetic ShapeNet dataset and evaluated it on either synthetic or real data.
Our method can also be trained on real-data that consists of image pairs of the same object which vary in their viewpoints.
To this end, we use the recently proposed Objectron dataset~\cite{objectron2020}, and the Freiburg cars dataset~\cite{sedaghat2015unsupervised}. 
For Objectron we train on the chair category, as it is present in ShapeNet and contains sufficiently diverse high-quality images in contrast to the other categories where images are blurry or there are too few videos.
While ViewNet does not require segmentation masks at test time, it does require segmented objects as the target for training.

\noindent{\bf Objectron.} 
We first randomly sample ten frames per videos, and obtain foreground masks using two different semantic segmentation methods:  DeepLabV3~\cite{chen2017rethinking}, trained on COCO~\cite{lin2014microsoft} ground-truth segmentation masks and a weakly supervised method~\cite{araslanov2020single}, trained on Objectron frames using only \emph{image-level} labels.
We start from a model pretrained on ShapeNet to prevent overfitting on the relatively low amount of instances from Objectron.
ViewNet without cycles obtains 91\% and 89\% accuracy with 8.8$^o$ and 10.1$^o$ median error on PASCAL3D+ cars for the supervised and weakly-supervised segmentation settings respectively. 
This is a significant improvement from the 83\% accuracy obtained by using only the ShapeNet trained model.

\noindent{\bf Freiburg Cars.}  
As the dataset only contains 48 videos, we use all frames, \ie between 120 and 130 per instance. 
We also use segmentation masks obtained from a pre-trained supervised Mask R-CNN model~\cite{he2017mask}. 
Results are shown in Table~\ref{tab:FCars}. ViewNet obtains stronger results than the unsupervised approach of~\cite{novotny2017learning}. 
Adding our cycle loss does not improve performances as real cars exhibit specular reflections that ViewNet is unable to reproduce. 
As~\cite{novotny2017learning} does not report accuracy, we estimated it to be 50\%, which correspond to a median error of 30$^o$.

\subsection{Limitations} 
ViewNet requires foreground masks at training time as the model is unable to extract background information from the appearance image. 
In experiments on real data, we use pre-trained segmentation models~\cite{chen2017rethinking, araslanov2020single, he2017mask} to estimate these masks.
However, it is important to note that the viewpoint estimator can be applied to unsegmented images at test time.
Our method also relies on having image pairs during training in order to disentangle viewpoint and object appearance, which limits its real-world application to video datasets.
Finally, we assume that object appearance is independent from the viewpoint, but this assumption is often violated by non-Lambertian surfaces, \eg cars. 
We qualitatively analyze some failure cases of our method in the supplementary material.

\begin{table}[t]
\small
\setlength{\tabcolsep}{3pt}
  \centering
  \begin{tabular}{lcccc}
    \toprule
                    & VGG view &\multicolumn{1}{p{40pt}}{VpDR-Net + FrC~\cite{novotny2017learning}} & ViewNet & \multicolumn{1}{p{35pt}}{ViewNet + cycle} \\
   \midrule
     Acc (\%)       &       56 &                                $\sim$ 50 & \B   61 &   59\\
   \midrule
     Med err ($^o$) &     25.8 &                                     29.6 & \B 16.1 & 19.1\\

  \bottomrule
  \end{tabular}
  \vspace{-5pt}
  \caption{Comparison of models trained on Freiburg Cars and evaluated on PASCAL3D+.}
  \vspace{-5pt}
  \label{tab:FCars}
\end{table}

%% file: conclusion.tex
% !TEX root = main.tex

We presented ViewNet, a self-supervised approach for learning object viewpoint estimation from image pairs.
By ensuring geometric consistency during generation, we can accurately synthesize new views from objects and use them to refine our network predictions, outperforming current approaches on both synthetic and real datasets.
Finally, we highlighted evaluation issues on the commonly used PASCAL3D+ dataset. 
We demonstrate that there are significant biases in the dataset and even simple baseline methods can perform well, suggesting a need for new benchmarks with more varied 3D poses.

%% file: supplementary.tex
In this supplementary document we provide additional results. In \cref{sec:debias} we evaluate ViewNet on PASCAL3D+ using an improved metric we have have created, then we provide a qualitative comparison of synthesized images from different models in \cref{sec:abl}, and finally in \cref{sec:impl} we describe additional implementation details used in our experiments.

\section{Debiasing Viewpoint Evaluation}
\label{sec:debias}

\begin{table*}[!b]
\small

  \centering
  \begin{tabular}{lcccccccccccc}
    \toprule
                                 & aeroplane &   bike &   boat & bottle &    bus &    car &  chair &  table &  mbike &   sofa &  train & tv \\
    Confidence index               &      1 &      1 &      1 &    .17 &     .5 &      1 &     .5 &    .67 &    .92 &    .42 &    .33 &    .33 \\
   \midrule
    Constant                       &     30 &     20 &     21 &     48 &     30 &     19 &     21 &     31 &     22 &     32 &     34 &     49 \\
    VGG view*                      &     49 &     58 &     33 &     47 &     32 &     56 &     37 &     37 &     32 &     40 &     36 &     60 \\
    ViewNet*                       &     60 &     67 &     19 &     48 &     33 &     84 &     49 &     36 &     64 &     57 &     31 &     50 \\
    ViewNet                        &     61 &     72 &     20 &     47 &     33 &     86 &     78 &     40 &     69 &     86 &     27 &     55 \\
    ViewNet + cycle *              &     53 &     62 &     21 &     48 &     29 &     85 &     55 &     34 &     67 &     50 &     34 &     50 \\
    ViewNet + cycle                &     62 &     71 &     21 &     54 &     30 &     85 &     61 &     37 &     76 &     76 &     43 &     58 \\
  
  \bottomrule
  \end{tabular}
  \caption{Discretized Viewpoint Accuracy. Entries followed by a star (*) use the linear regression alignment procedure described in the main paper, and those without use a single global alignment.}
  \label{tab:intervaled_accuracy}
\end{table*}

The standard evaluation used with the PASCAL3D+ dataset involves computing an average of the viewpoint prediction accuracy across the entire  evaluation set and the median error in degrees for a given object class.
As discussed in the main text, many categories in the PASCAL3D+ dataset are strongly biased as test images are taken from a limited number of viewpoints.
We illustrated this problem by creating a simple baseline, ``Constant'', that outputs the average viewpoint in the training set. 
We Table 3 in the main paper we showed that this achieves surprisingly strong  performance in some categories \eg ``bottle'' (96\%), ``bus'' (79\%), ``sofa'' (81\%),``train'' (95\%), and ``tv'' (89\%). 
This is explained by the highly concentrated number of viewpoints in the data.
Note that this issue has also been reported in Figure 2 of the original PASCAL3D+ paper.

To address the issue caused by this viewpoint bias, we propose a more balanced evaluation by introducing a new metric, Discretized Viewpoint Accuracy (DVA). 
To this end, we split the ground-truth viewpoints in the evaluation set into bins, each spanning $30^o$ azimuth-wise and compute standard viewpoint accuracy for each bin, before averaging the results. 
This ensures highly populated viewpoints do not cause performances to be overestimated and requires a model to perform well over not only a single subset but all subsets to reach high performances. 
Clearly, when there are no samples belong to a bin, it is not possible to measure the performance in this interval.
Hence we omit empty bins in the evaluation.
Note that in the extreme case where all samples belong to a single bin, DVA is equal to the standard viewpoint accuracy, limiting its effect as an unbiased metric.
To account for this, we also compute an auxiliary dataset statistic called Confidence Index. 
It is defined as the fraction of bins having more than 10 samples. 
Hence, categories with low confidence index should be interpreted cautiously.
In Table~\ref{tab:intervaled_accuracy} we present results using our DVA metric.

\section{Qualitative results}
\label{sec:abl}

\subsection{Synthetic images}
In addition to the quantitative ablation study in Table~2 in the main paper, here we provide a supporting qualitative analysis.
In particular, we illustrate image reconstructions for ViewNet as well as three different ablation experiments on the ShapeNet dataset:
\begin{enumerate}
   \setlength\itemsep{-4pt}
    \item No Shape Prior: removing the Gaussian shape prior from the decode, that is, trying to learn $Q$ directly.
    \item Encoder-Decoder: Using a regular encoder-decoder architecture instead of decoding a canonical code adapted with adaptive instance normalization.
    \item HoloGAN-Decoder: Using extra convolution layers after the rotation and projection, as per HoloGAN and SSV.
\end{enumerate}

Reconstructions for different variants of ViewNet  are shown in Figure~\ref{fig:ablation}. These were obtained by feeding the leftmost image to the appearance network, then sampling the viewpoint space at regular azimuth and decoding the representation along those viewpoints. We note the difference in canonical viewpoints adopted by each model, as the first image in each series corresponds to different viewpoints. Interestingly, all models have learned to remove the piece of ground that appears under the car in the appearance image (visible as a plane that can be seen when zooming in), most likely because this is an uncommon feature in the dataset.

The black spots that appear around the object (Figure~\ref{fig:shape}) indicate that the model fails to learn the proper shape of the object and considers the background as part of the object. An extreme case of this is when the model does not learn shape at all and tries to ``paint'' the object on the volume, such as chairs in Figure~\ref{fig:shape}. 
This emphasizes the importance of our shape prior. 

Although their performance is similar, the views generated using an Encoder-Decoder architecture in Figure~\ref{fig:AdaIN} are not as faithful to the original object compared to using a generator with adaptive instance normalization.
More precisely, the objects tend to be closer to the average object in the category, with the aeroplane being gray, as most aircraft tend to be, and the chair seat being square instead of round.
The lack of fidelity in the reconstructions are apparent when comparing to ViewNet in Figure~\ref{fig:standard}. 

For the HoloGAN-style generation (Figure~\ref{fig:HoloGAN}), there is no distinction between object and background as additional layers are used after the projection stage, translating in a black background for all generated images. 
We note that the geometry is not preserved, and we see poor consistency when traversing the viewpoint space, which is most apparent in the case of the airplane.

\subsection{Real images}
\begin{figure*}[t]
    \centering
    \begin{subfigure}{.33\linewidth}
        \includegraphics[width=\linewidth]{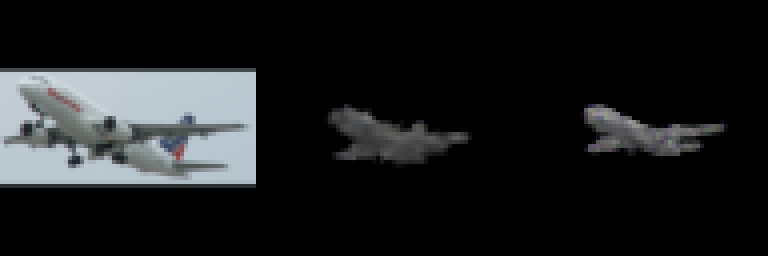}
        \includegraphics[width=\linewidth]{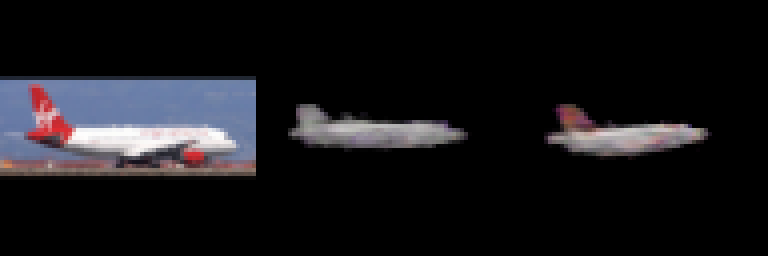}
        \includegraphics[width=\linewidth]{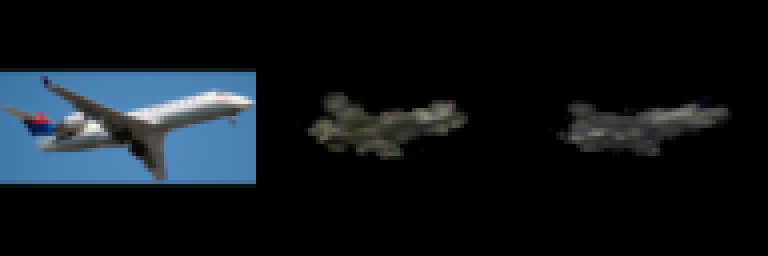}
        \caption{Airplanes}
    \end{subfigure}
    \begin{subfigure}{.33\linewidth}
        \includegraphics[width=\linewidth]{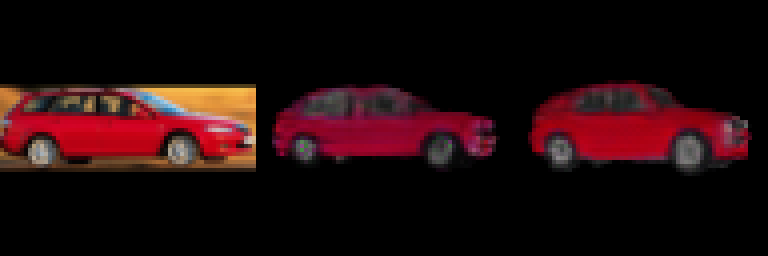}
        \includegraphics[width=\linewidth]{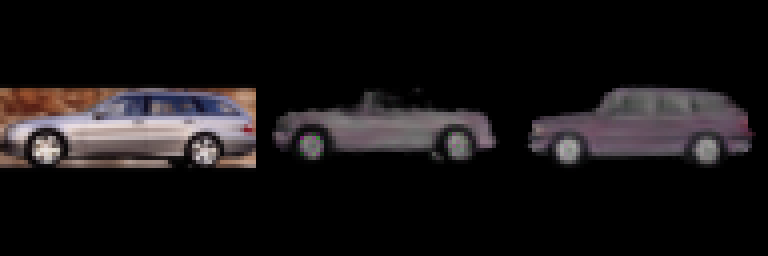}
        \includegraphics[width=\linewidth]{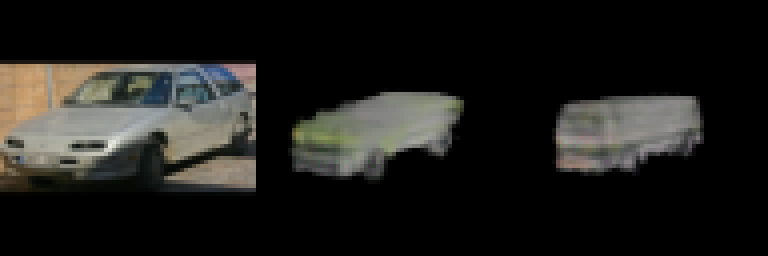}
        \caption{Cars}
    \end{subfigure}
    \begin{subfigure}{.33\linewidth}
        \includegraphics[width=\linewidth]{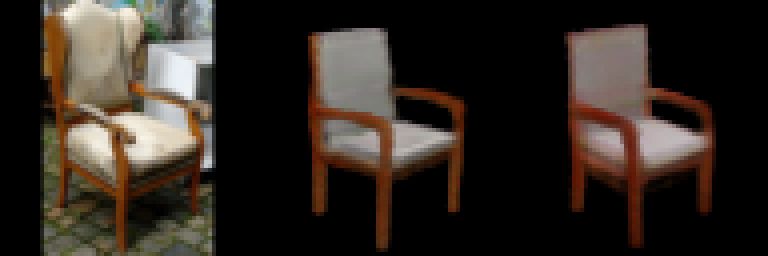}
        \includegraphics[width=\linewidth]{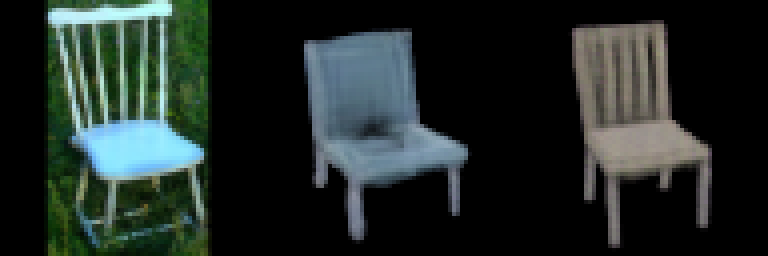}
        \includegraphics[width=\linewidth]{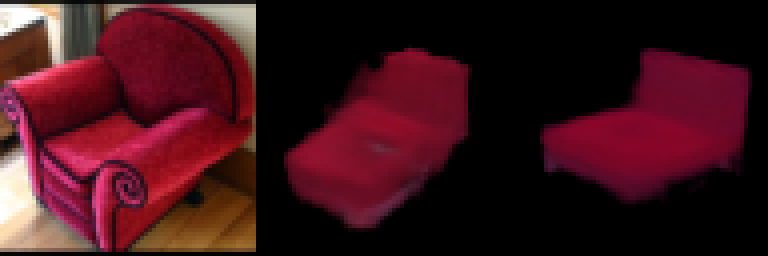}
        \caption{Chairs}
    \end{subfigure}
    \caption{Reconstructions of real images from PASCAL3D+. 
    For each of the three object categories, the left image is the original, the middle image is the standard ViewNet reconstruction, and the right image is the ViewNet + cycle reconstruction. 
    }
    \label{fig:recons}
\end{figure*}

\begin{figure*}
    \centering
    \begin{subfigure}{\textwidth}
        \centering
        \includegraphics[width=\linewidth]{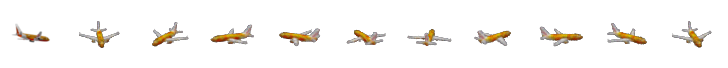}\\
        \includegraphics[width=\linewidth]{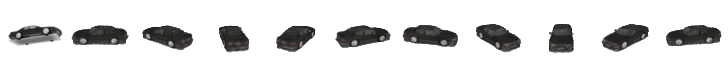}\\
        \includegraphics[width=\linewidth]{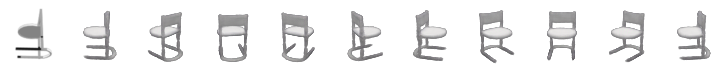}\\
        \caption{ViewNet}
        \label{fig:standard}
    \end{subfigure}
    
    \begin{subfigure}{\textwidth}
        \centering
        \includegraphics[width=\linewidth]{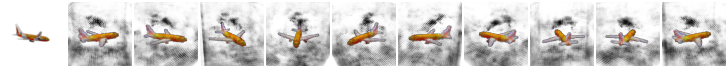}\\
        \includegraphics[width=\linewidth]{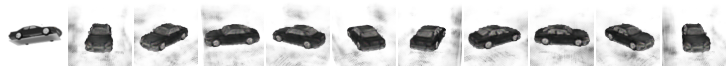}\\
        \includegraphics[width=\linewidth]{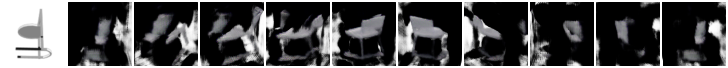}\\
        \caption{No Shape Prior}
        \label{fig:shape}
    \end{subfigure}
    
    \begin{subfigure}{\textwidth}
        \centering
        \includegraphics[width=\linewidth]{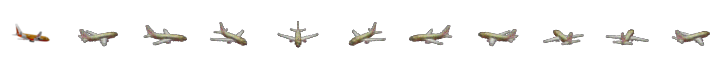}\\
        \includegraphics[width=\linewidth]{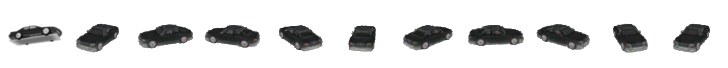}\\
        \includegraphics[width=\linewidth]{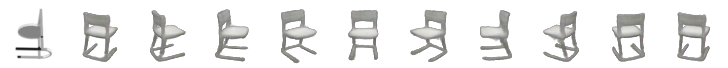}\\
        \caption{Encoder-Decoder}
        \label{fig:AdaIN}
    \end{subfigure}
    
    \begin{subfigure}{\textwidth}
        \centering
        \includegraphics[width=\linewidth]{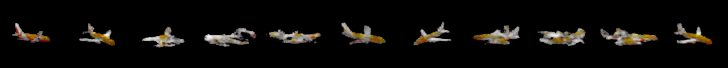}\\
        \includegraphics[width=\linewidth]{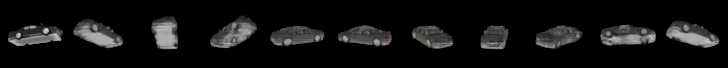}\\
        \includegraphics[width=\linewidth]{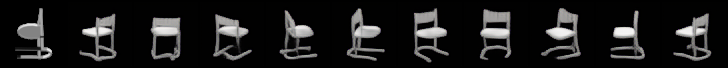}\\
        \caption{HoloGAN-Decoder}
        \label{fig:HoloGAN}
    \end{subfigure}
        
    \caption{Generated views at constant elevation for ablated models. The left most image provides the object appearance.}
    \label{fig:ablation}
\end{figure*}

We also show reconstructions of real objects from unobserved PASCAL3D+ images based on ShapeNet-trained models, for both ViewNet and ViewNet + cycle, in Figure~\ref{fig:recons}. 
This is illustrated by the fact that the rendered image from the model has the same pose as the input image. 
Note that background is not part of the reconstruction, as the models were trained to reconstruct objects only \cf section~4.3. 
The middle row of each category shows in particular how adding cycles has helped the model gain better understanding of the object, leading to more faithful reconstruction, \eg in the red tail of the plane or the vertical bars in the chair. The bottom row show failure cases where models failed to capture the object appearance. We observe that even when this is the case, the viewpoint is still correctly predicted.

\begin{figure*}
    \centering
    \includegraphics[width=.13\textwidth]{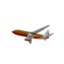}
    \includegraphics[width=.13\textwidth]{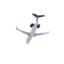}
    \includegraphics[width=.13\textwidth]{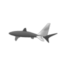}
    \includegraphics[width=.13\textwidth]{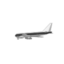}
    \includegraphics[width=.13\textwidth]{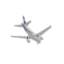}
    \includegraphics[width=.13\textwidth]{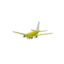}
    \includegraphics[width=.13\textwidth]{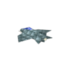}\\
    \includegraphics[width=.13\textwidth]{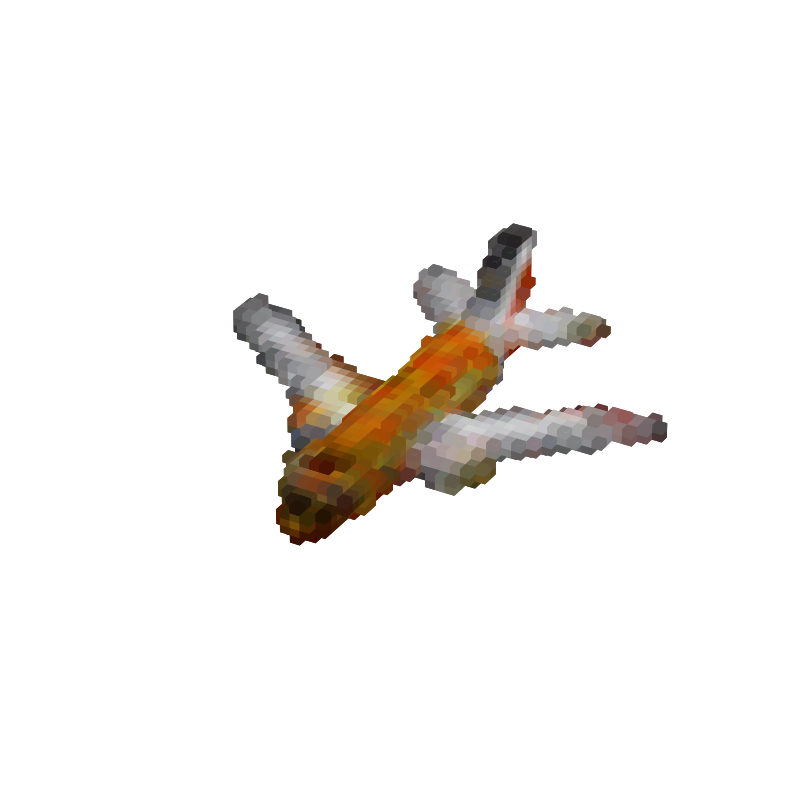}
    \includegraphics[width=.13\textwidth]{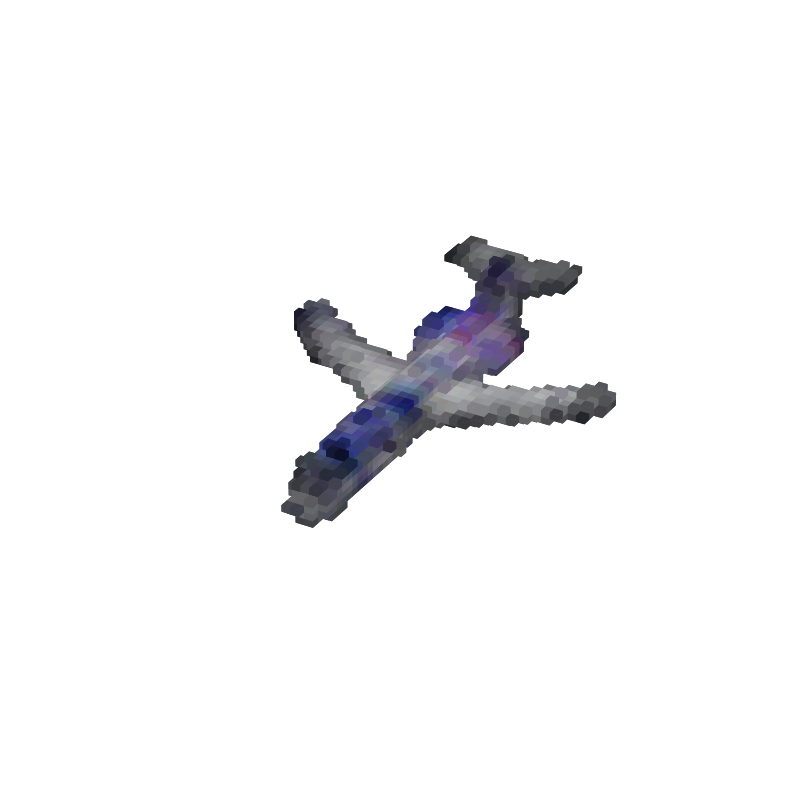}
    \includegraphics[width=.13\textwidth]{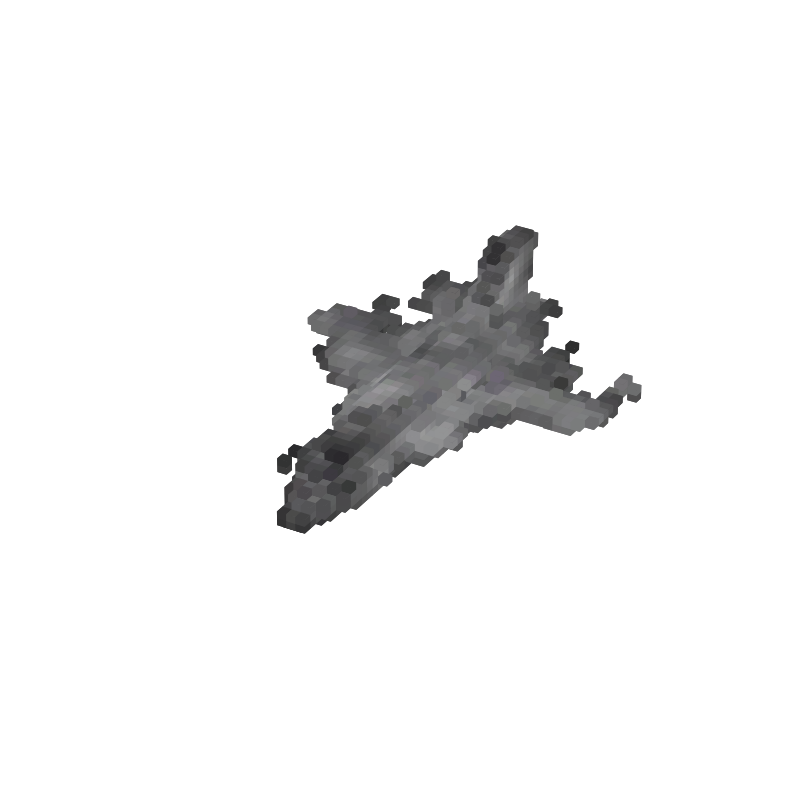}
    \includegraphics[width=.13\textwidth]{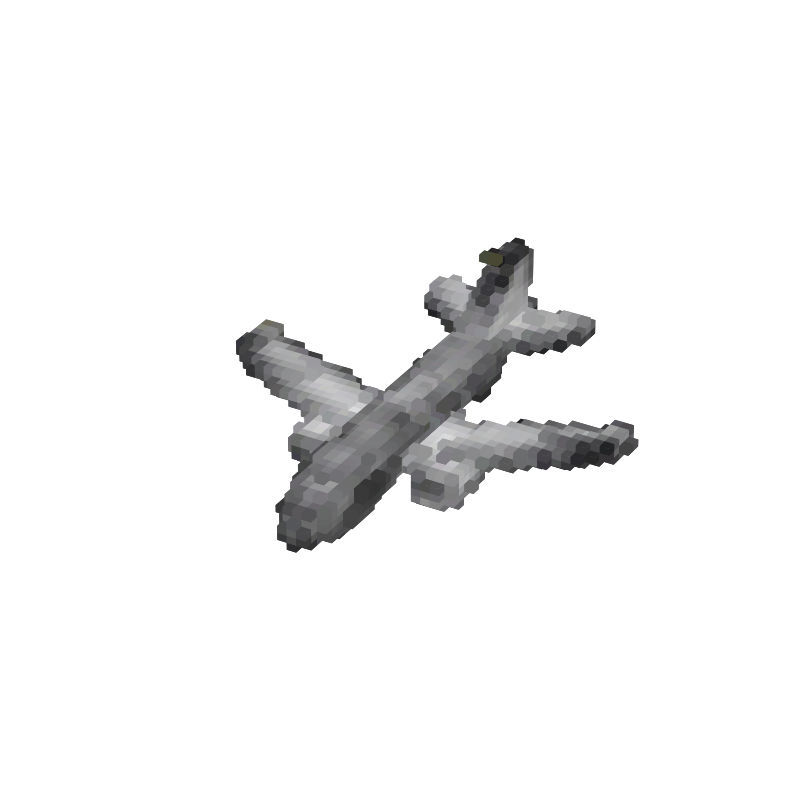}
    \includegraphics[width=.13\textwidth]{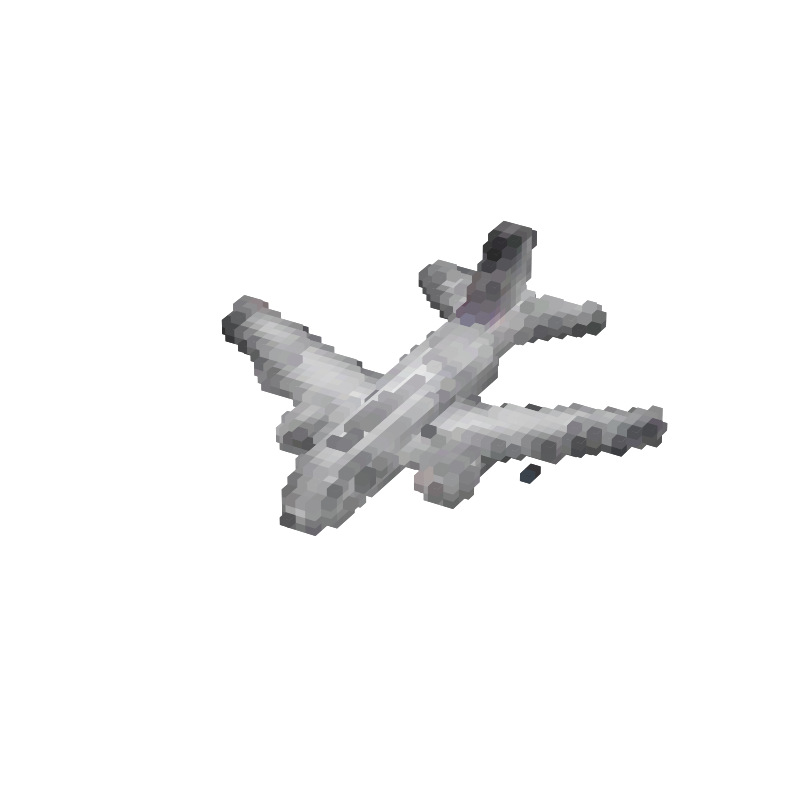}
    \includegraphics[width=.13\textwidth]{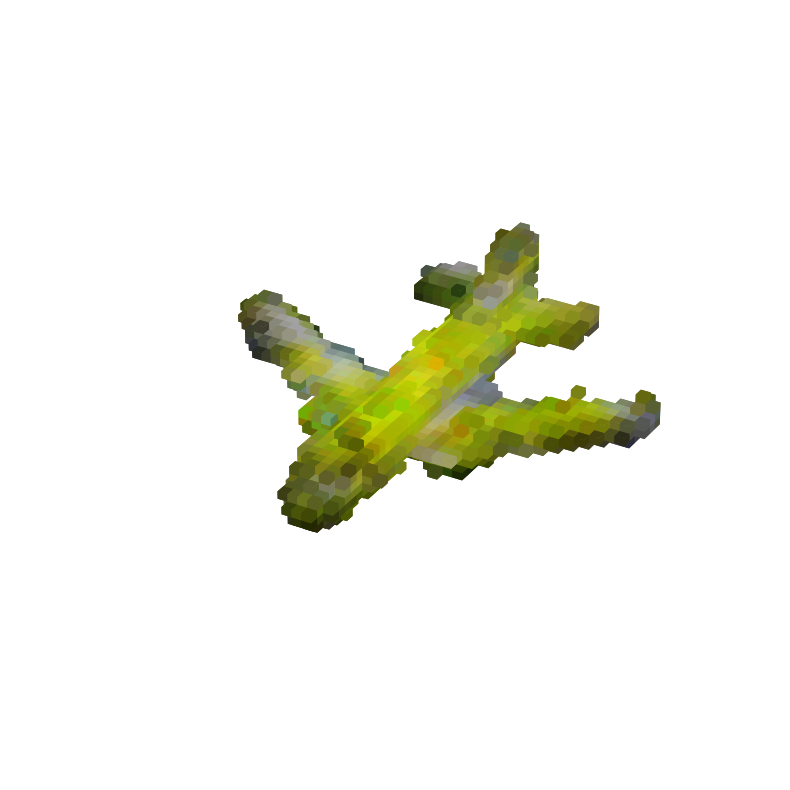}
    \includegraphics[width=.13\textwidth]{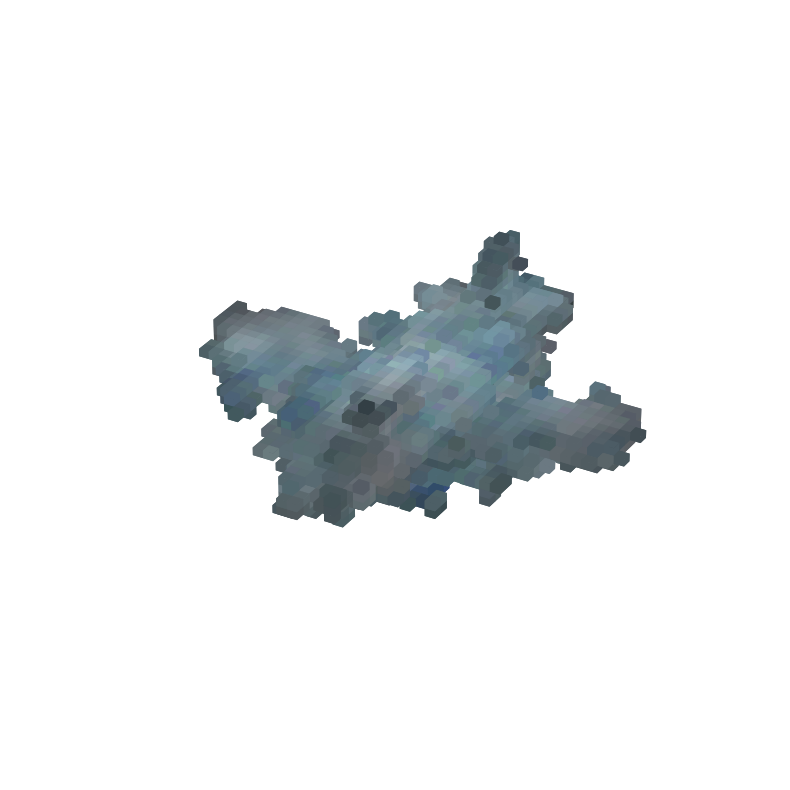}\\
    \includegraphics[width=.13\textwidth]{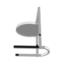}
    \includegraphics[width=.13\textwidth]{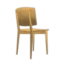}
    \includegraphics[width=.13\textwidth]{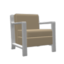}
    \includegraphics[width=.13\textwidth]{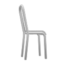}
    \includegraphics[width=.13\textwidth]{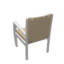}
    \includegraphics[width=.13\textwidth]{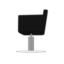}
    \includegraphics[width=.13\textwidth]{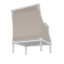}\\
    \includegraphics[width=.13\textwidth]{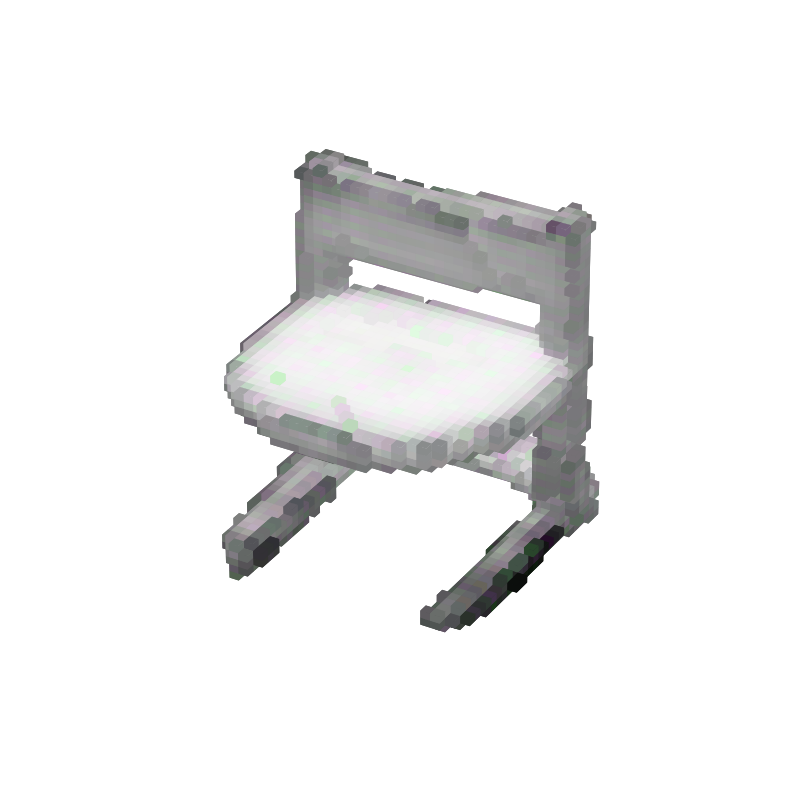}
    \includegraphics[width=.13\textwidth]{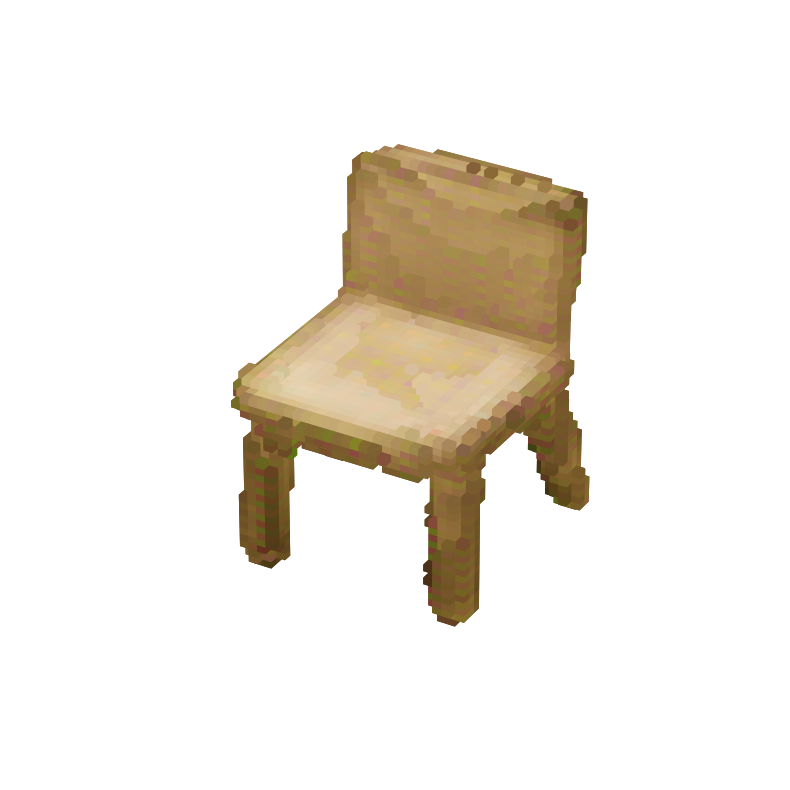}
    \includegraphics[width=.13\textwidth]{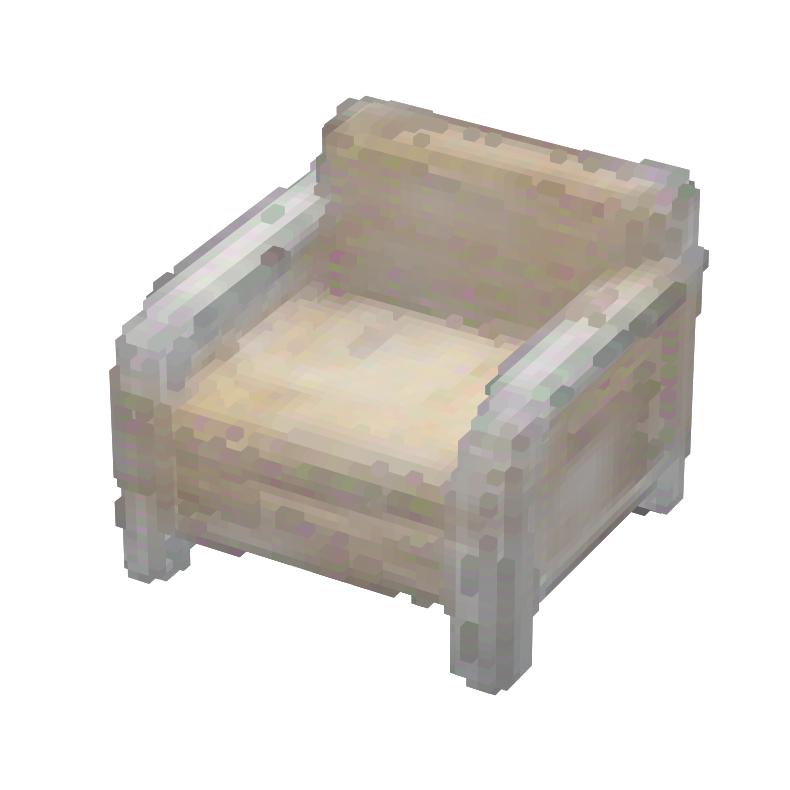}
    \includegraphics[width=.13\textwidth]{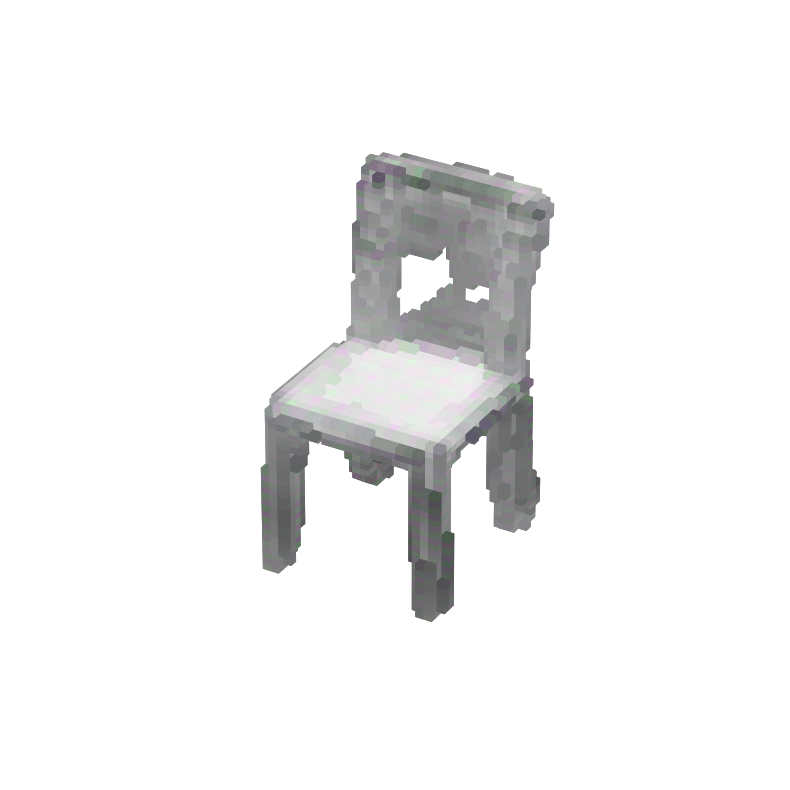}
    \includegraphics[width=.13\textwidth]{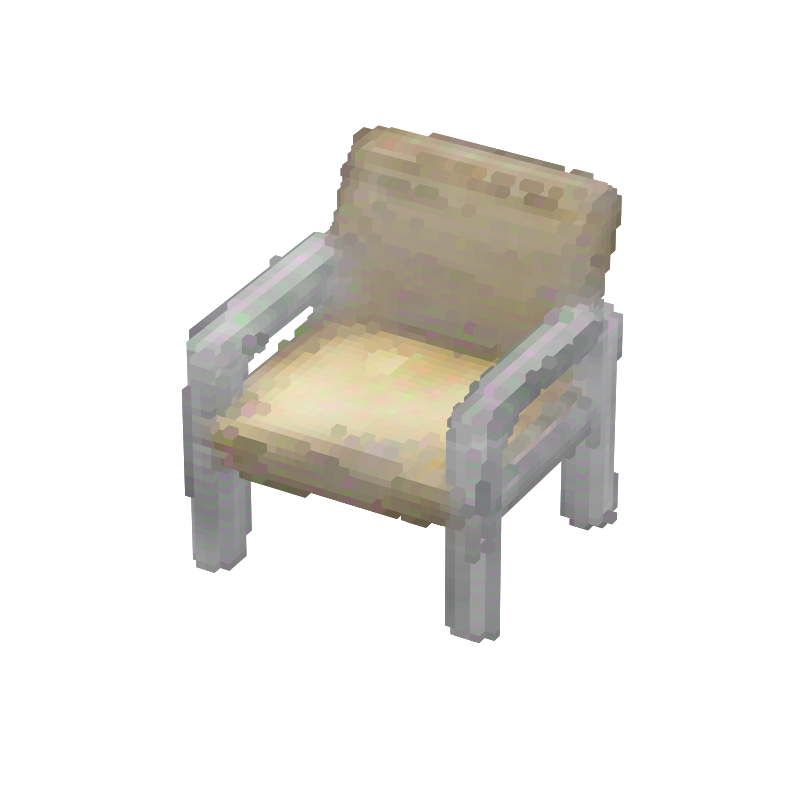}
    \includegraphics[width=.13\textwidth]{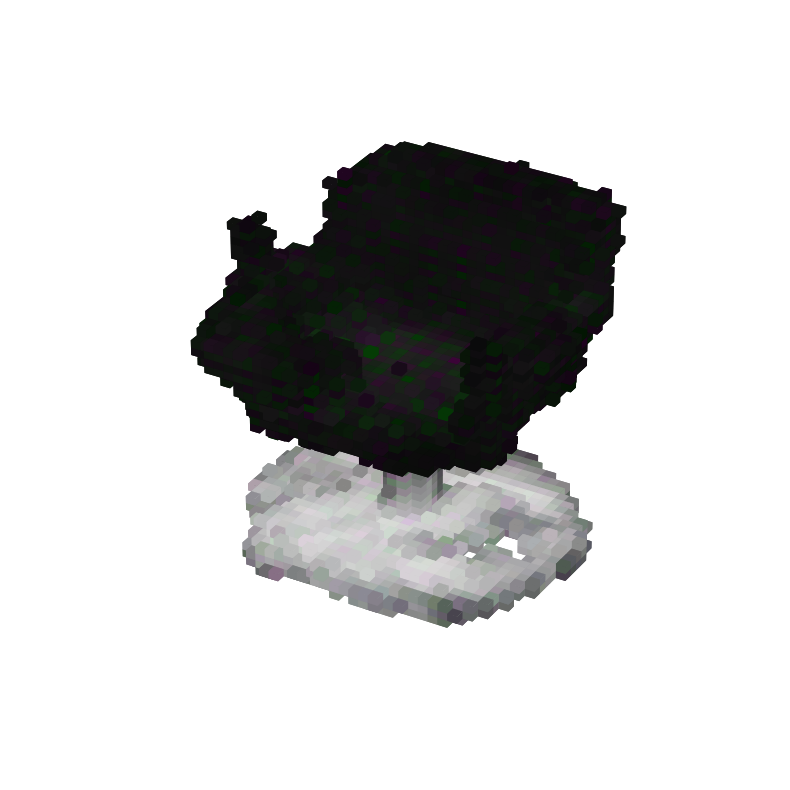}
    \includegraphics[width=.13\textwidth]{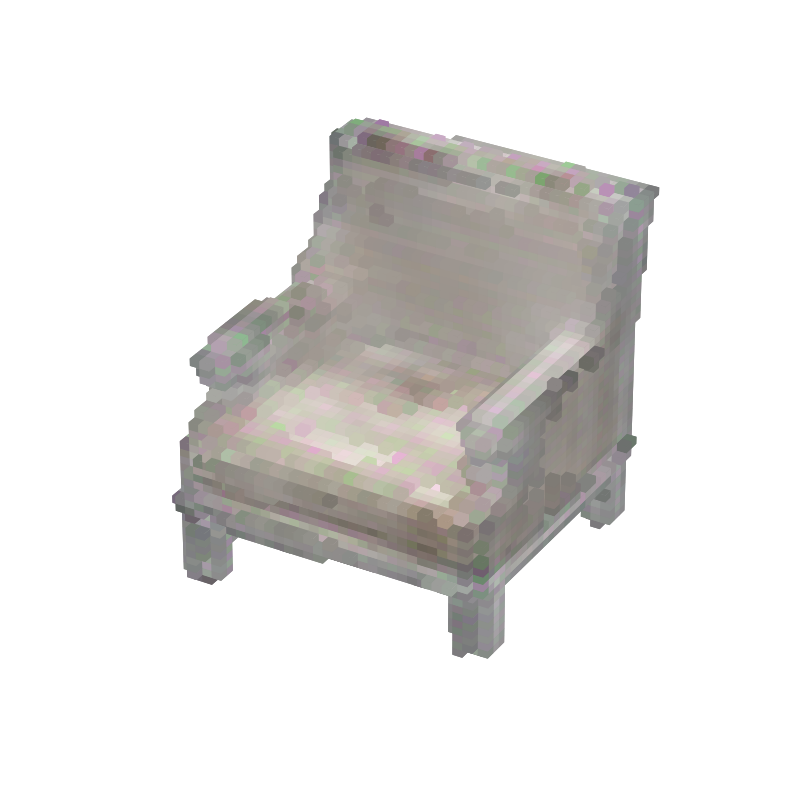}\\
    \includegraphics[width=.13\textwidth]{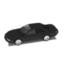}
    \includegraphics[width=.13\textwidth]{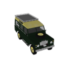}
    \includegraphics[width=.13\textwidth]{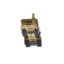}
    \includegraphics[width=.13\textwidth]{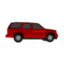}
    \includegraphics[width=.13\textwidth]{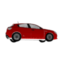}
    \includegraphics[width=.13\textwidth]{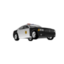}
    \includegraphics[width=.13\textwidth]{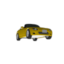}\\
    \includegraphics[width=.13\textwidth]{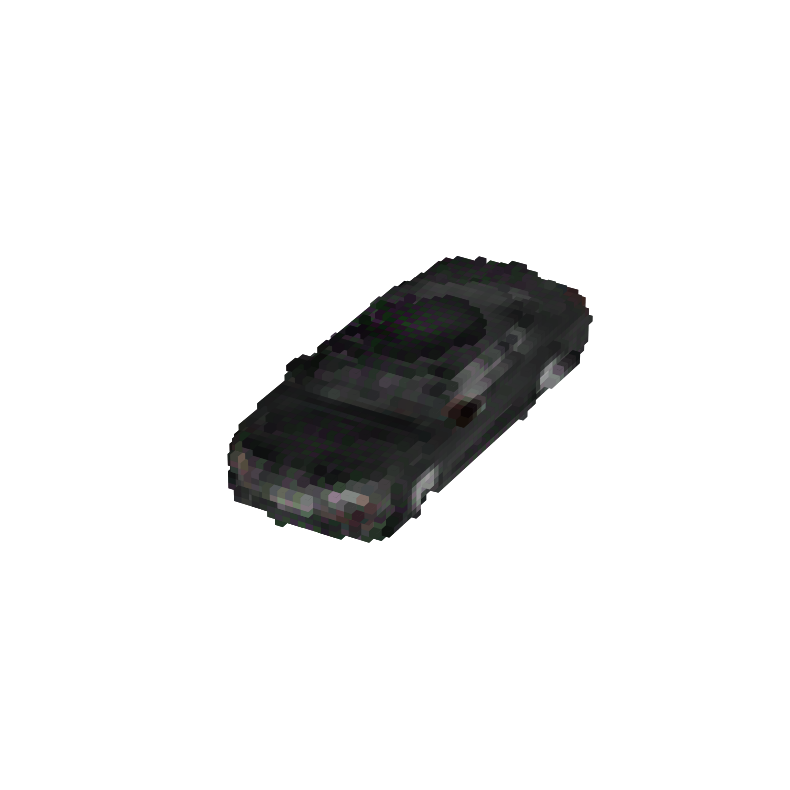}
    \includegraphics[width=.13\textwidth]{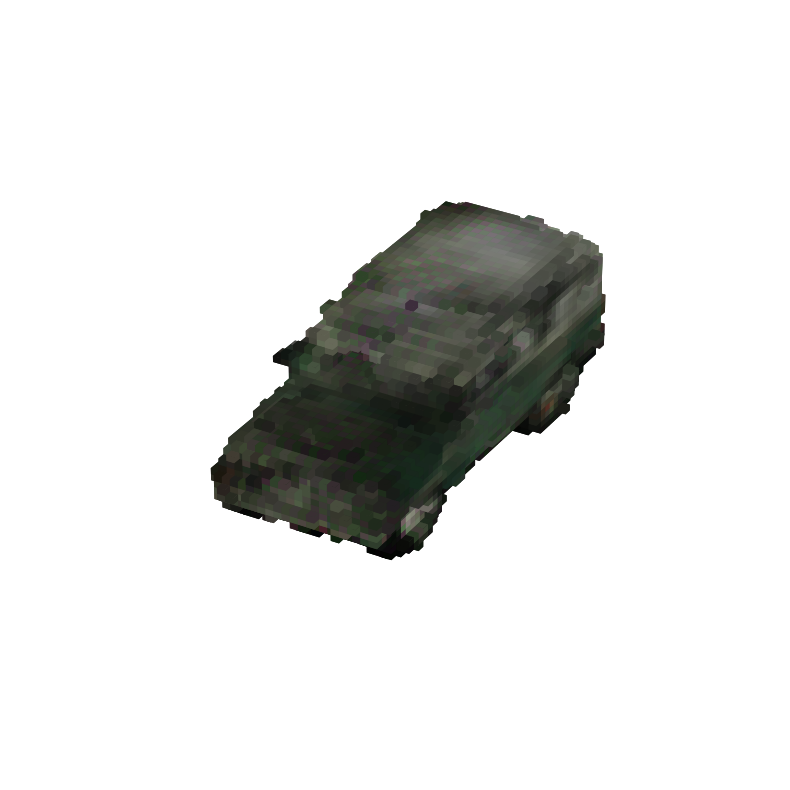}
    \includegraphics[width=.13\textwidth]{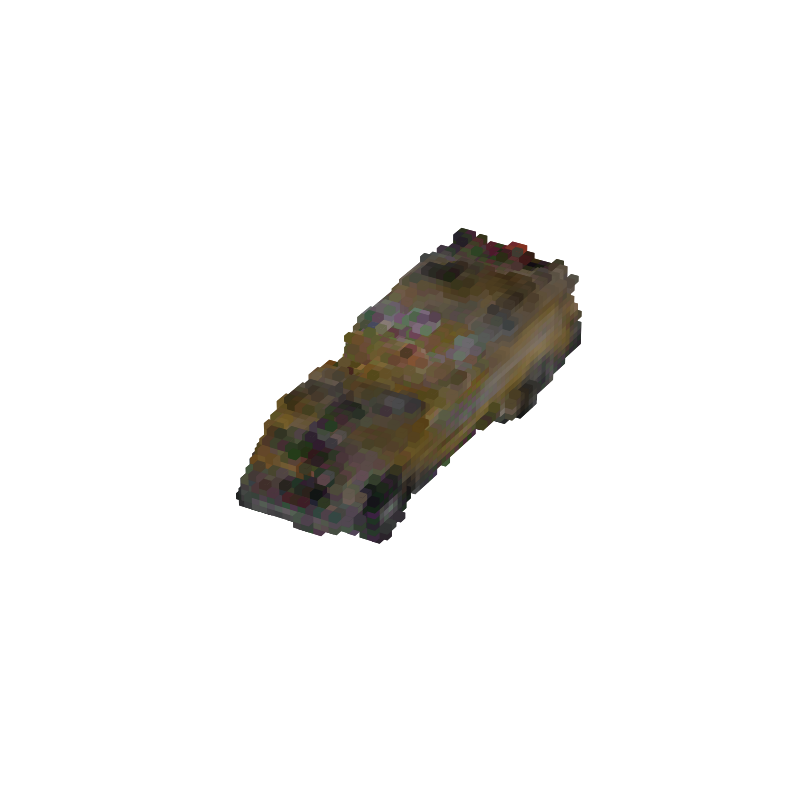}
    \includegraphics[width=.13\textwidth]{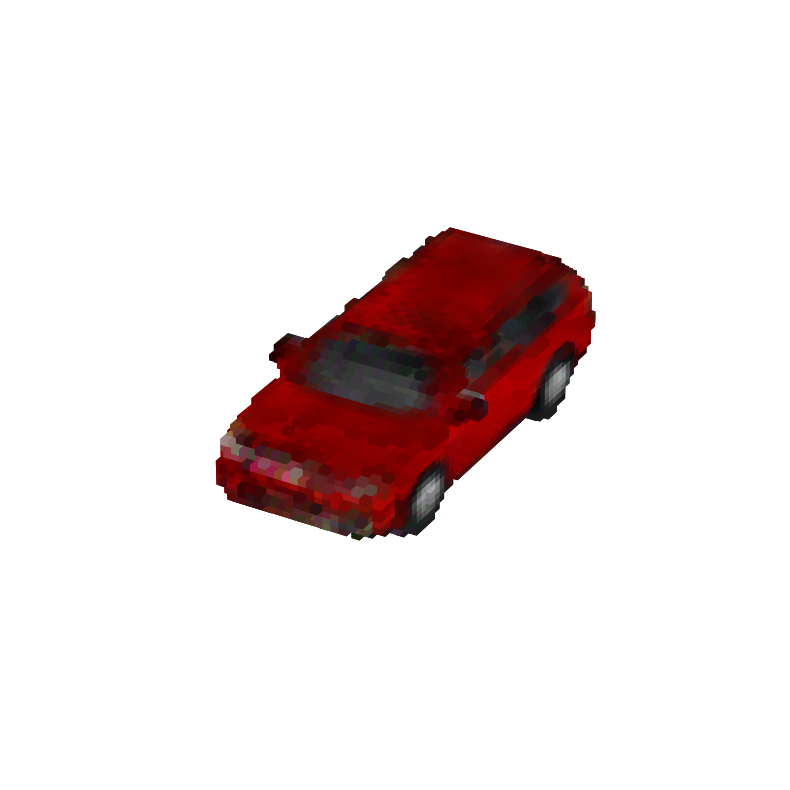}
    \includegraphics[width=.13\textwidth]{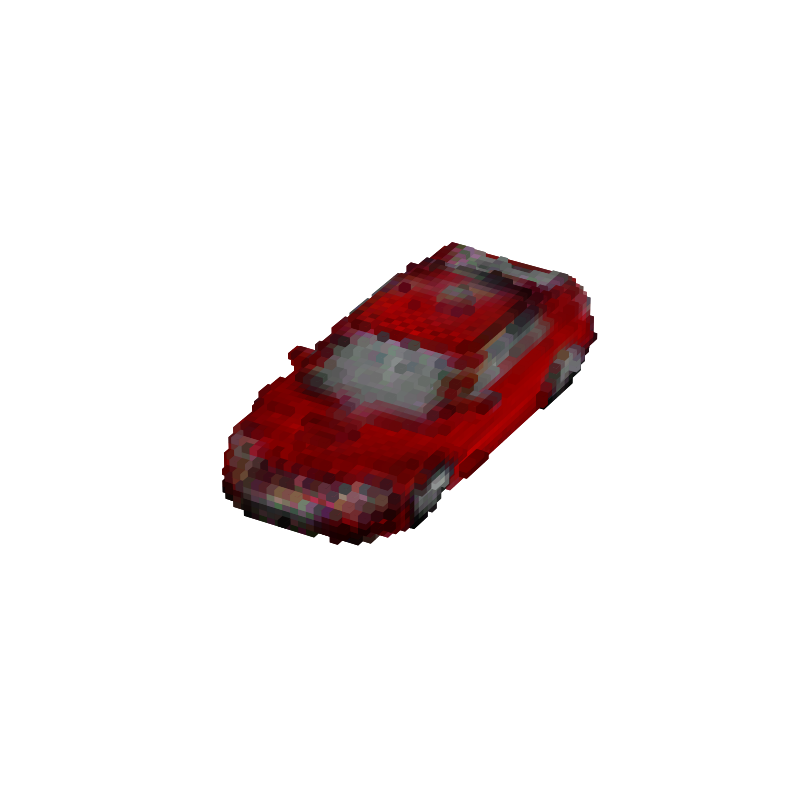}
    \includegraphics[width=.13\textwidth]{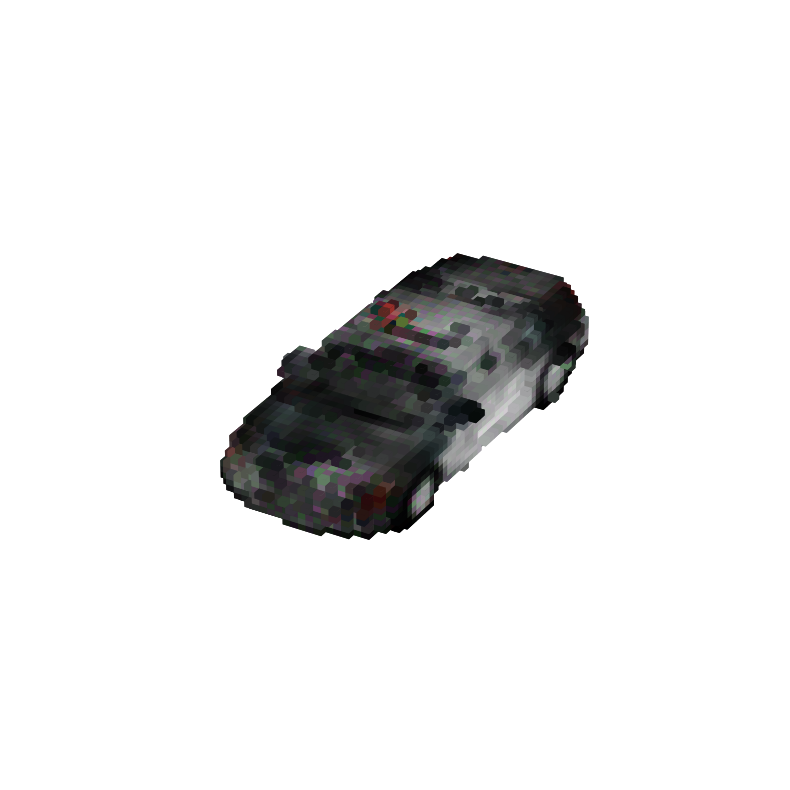}
    \includegraphics[width=.13\textwidth]{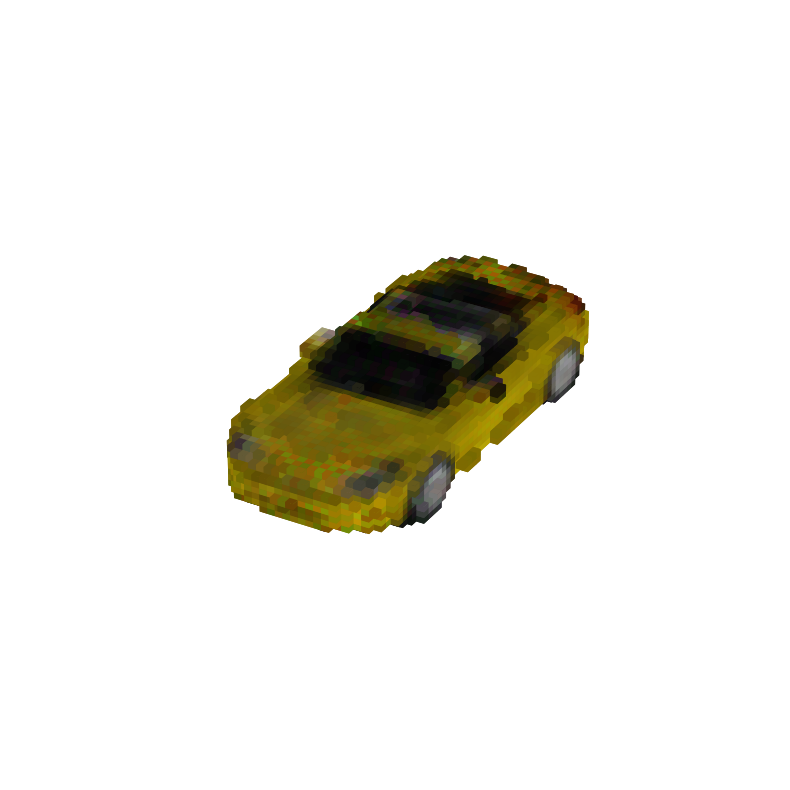}\\
    \caption{Visualization of 3D models learned from different appearance images. The top images is fed to the appearance network, and the bottom bottom one is the output of the decoder is shown in the canonical pose}
    \label{fig:3D_voxels}
\end{figure*}

\section{Network architecture}
\label{sec:impl}
Finally, we describe the network architectures used. 
Both $f^v$ and $f^a$ use the encoder architecture illustrated in Table~\ref{tab:encoder} with output sizes 3 and 256 respectively, while $f^d$ uses the generator architecture in Table~\ref{tab:decoder}.

\begin{table*}[ht]
    \begin{minipage}{.5\linewidth}
        \footnotesize
        \centering
        \begin{tabular}{lccc}
            \toprule
            Layer   & \# channels& Kernel& Stride\\
            \midrule
            Conv2D  & 64    &3x3   & 2\\
            BatchNorm\\
            ReLU\\
            \midrule
            Conv2D  & 128    &3x3   & 2\\
            BatchNorm\\
            ReLU\\
            \midrule
            Conv2D  & 256    &3x3   & 2\\
            BatchNorm\\
            ReLU\\
            \midrule
            Conv2D  & 512    &3x3   & 2\\
            BatchNorm\\
            ReLU\\
            \midrule
            Conv2D  & 512    &3x3   & 2\\
            BatchNorm\\
            ReLU\\
            \midrule
            Conv2D  & 512    &2x2   & 1\\
            BatchNorm\\
            ReLU\\
            \midrule
            Conv2D  & variable    &1x1   & 1\\
            BatchNorm\\
            ReLU\\
            
            \bottomrule
        \end{tabular}
        \caption{Encoder architecture.}
        \label{tab:encoder}
        
    \end{minipage}
    \hfill
    \begin{minipage}{.5\linewidth}
        \footnotesize
        \centering
        \begin{tabular}{lccc}
            \toprule
            Layer   & \# channels& Kernel& Stride\\
            \midrule
            ConvTranspose3D  & 512    &4x4   & 1\\
            AdaptiveIN\\
            ReLU\\
            
            \midrule
            ConvTranspose3D  & 128    &4x4   & 2\\
            AdaptiveIN\\
            ReLU\\
            
            \midrule
            ConvTranspose3D  & 128    &3x3   & 1\\
            AdaptiveIN\\
            ReLU\\
            
            \midrule
            ConvTranspose3D  & 128    &4x4   & 2\\
            AdaptiveIN\\
            ReLU\\
            
            \midrule
            ConvTranspose3D  & 128    &3x3   & 1\\
            AdaptiveIN\\
            ReLU\\
            
            \midrule
            ConvTranspose3D  & 16     &4x4   & 2\\
            AdaptiveIN\\
            ReLU\\
            
            \midrule
            ConvTranspose3D  & 16    &3x3   & 1\\
            AdaptiveIN\\
            ReLU\\
            
            \midrule
            ConvTranspose3D  & 4    &4x4   & 2\\
            AdaptiveIN\\
            ReLU\\

            \bottomrule
        \end{tabular}
        \caption{Generator architecture.}
        \label{tab:decoder}
         
    \end{minipage}
\end{table*}